\documentclass[10pt,twocolumn,letterpaper]{article}

\usepackage{cvpr}              % To produce the CAMERA-READY version
\usepackage{booktabs}
\usepackage{multirow}
\usepackage{graphicx}
\usepackage{booktabs,adjustbox,xcolor,colortbl}
\usepackage[T1]{fontenc}
\usepackage[utf8]{inputenc}
\usepackage{microtype,tabularx,array,booktabs}
\usepackage{tabularx}
\usepackage{array}
\usepackage{amsmath,amssymb}
\usepackage{microtype}
\usepackage{xcolor}
\usepackage{enumitem}
\newcolumntype{Y}{>{\raggedright\arraybackslash}X}

\newcommand{\code}[1]{\texttt{#1}}

\usepackage{framed,needspace,tikz}
\definecolor{cvprblue}{rgb}{0.21,0.49,0.74}
\definecolor{cvprblue}{rgb}{0.21,0.49,0.74}
\definecolor{aqaInk}{HTML}{235D55}
\definecolor{aqaTint}{HTML}{E8F2EF}
\definecolor{aqaGray}{HTML}{F1F3F4}
\usepackage[pagebackref,breaklinks,colorlinks,allcolors=cvprblue]{hyperref}

\def\confName{CVPR}
\def\confYear{2027}

\title{StreamCoach: Proactive Sports Coaching via Continuous Visual Reasoning}

\author{Xiaoye Liang$^{1,2}\thanks{Equal contribution}$ \quad Yiming Huang$^{2,3*}$ \quad Jialu Wu$^{4}$ \quad Yushe Cao$^{5}$ \quad Ruxiang Duan$^{6}$ \quad Yi Sun$^{2,7}$ \quad Tie Liu$^{1}$ \\ Lai Jiang$^{1}$ \quad Mai Xu$^{1}$ \quad Shikun Feng$^{2,8}$\thanks{Corresponding author}  \quad Yiheng Zhu$^{2,8\dagger}$ \\
$^{1}$ Beihang University \quad
$^{2}$ Zhongguancun Academy \quad
$^{3}$ Nankai University \quad
$^{4}$ Zhejiang University \\
$^{5}$ Tsinghua University \quad
$^{6}$ South China University of Technology  \\
$^{7}$ Institute of Automation, Chinese Academy of Science  \\
$^{8}$ Zhongguancun Institute of Artificial Intelligence
\\
{\tt\small \{fengshikun, zhuyiheng\}@zgci.ac.cn}
}

\begin{document}
\maketitle
\begin{abstract}
In fitness and sports training, timely action quality assessment (AQA) is essential for learners to identify and correct technical errors during movement execution. However, most AQA methods assess completed performances, limiting their ability to provide timely feedback. Streaming methods alleviate this latency, but limited reasoning hinders their use of subtle movement cues and sport-specific knowledge. Moreover, their fragmented feedback focuses on one body part at a time, overlooking movement details of other body parts and transitions between action phases.
To address these limitations, we propose StreamCoach, a streaming AQA framework under the OnCueCoaching paradigm that couples continuous chain-of-thought (CoT) reasoning with proactive, evidence-gated coaching.
Technique-Aware Reasoning tracks body-part movements and action-phase transitions against sport-specific criteria, while Holistic Coaching integrates accumulated evidence to deliver timely, comprehensive feedback on strengths and technical errors.
To support training and benchmarking,  we construct EE4D-Stream from 1,213 sports video sequences in Ego-Exo4D through LLM-assisted reannotation and manual review of selected claims, pairing dense CoT annotations with evidence-gated speech content.
Building on this dataset, we train StreamCoach using Temporal–Semantic Alignment Reinforcement Learning to jointly optimize feedback timing and content.
We benchmark 11 offline and streaming baselines on EE4D-Stream, including three proprietary models. StreamCoach achieves state-of-the-art results across intervention timing, action assessment, and coaching quality, surpassing both GPT-6.1 Sol and Claude Opus 5.5.

\end{abstract} 
\section{Introduction}
In fitness and sports training, timely action quality assessment (AQA) helps learners correct technical errors during an action or before their next attempt.
For example, when a player pushes the ball too far ahead during a soccer slalom drill, immediate feedback can help them soften their touch before reaching the next pole. However,  many existing AQA methods require the full video for assessment, delaying corrective feedback and allowing errors to persist~\cite{pirsiavash2014assessing,parmar2017learning,tang2020uncertainty,yu2021group,xu2022finediving,zhou2026continual}. 
Therefore, real-time coaching calls for streaming AQA methods that assess technique as the action unfolds and provide targeted feedback once sufficient evidence accumulates.

Recent work explores assessment and task guidance in streaming video~\cite{panchal2024say,liu2026guideme,ji20263d}. For example, Panchal et al.~\cite{panchal2024say} introduced QEVD-Fit-Coach and Stream-VLM to learn when to intervene and what feedback to provide during exercise. 
Beyond fitness coaching, Liu et al.~\cite{liu2026guideme} introduced GuideMe, a multi-domain benchmark with annotations for next-step instructions, error detection, and corrective guidance in streaming video. 
Despite these advances, three limitations remain:
(1) \textbf{Limited reasoning}:  existing methods ~\cite{panchal2024say,ji20263d,liu2026guideme} struggle to integrate fine-grained movement cues with sport-specific knowledge and action intent. 
(2) \textbf{Fragmented feedback}: they ~\cite{panchal2024say,ji20263d,liu2026guideme} often identify only an overall issue or isolated error, without explaining what to retain, what to correct, and how to improve. 
(3) \textbf{Narrow data coverage}: existing long-form coaching datasets, e.g., QEVD-Fit-Coach~\cite{panchal2024say}, remain small and scenario-specific, restricting diversity and generalizability.

\begin{figure*}[tb]
  \centering
  \includegraphics[width=1\textwidth]{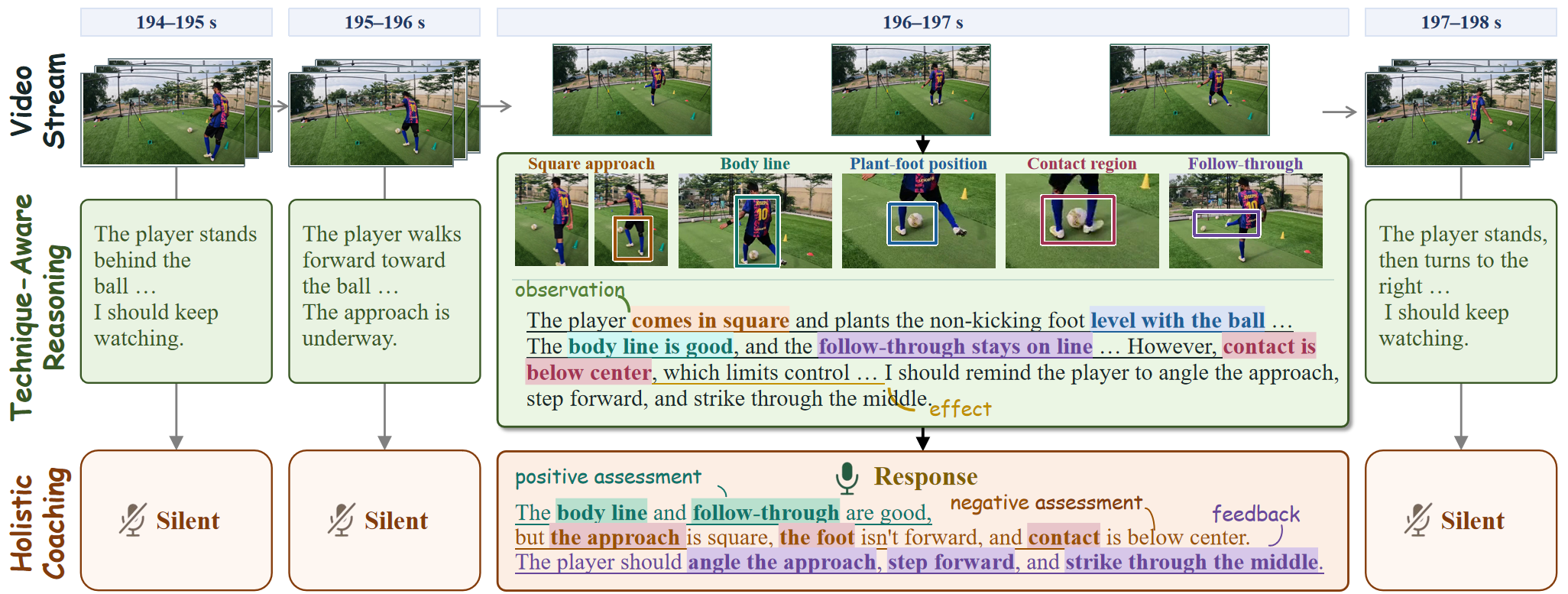}
  \caption{Illustration of the OnCueCoaching paradigm with a soccer sample. Technique-Aware Reasoning updates the assessment as visual evidence accumulates, while Holistic Coaching delivers feedback only when that evidence warrants intervention.}
  \label{fig:intro}
\end{figure*}

To address these limitations, we introduce OnCueCoaching, a streaming AQA paradigm that reasons over fine-grained movement cues as an action unfolds. 
As illustrated in Fig.~\ref{fig:intro}, OnCueCoaching performs Technique-Aware Reasoning at each streaming step and determines whether to provide Holistic Coaching. Specifically, Technique-Aware Reasoning integrates new observations of body-part motion and phase transitions with prior evidence to infer their implications for execution quality and update the assessment. Based on this assessment, Holistic Coaching triggers feedback only when the accumulated evidence warrants it. The feedback highlights strengths and pairs each identified error with a specific correction.
Building on this paradigm, we introduce StreamCoach, a streaming AQA framework with chain-of-thought (CoT) reasoning. We further introduce Temporal--Semantic Alignment Reinforcement Learning, which uses GRPO~\cite{shao2024deepseekmath} to jointly optimize feedback timing and content over full streaming trajectories.
To support training and benchmarking, we construct EE4D-Stream from 1,213 Ego-Exo4D video sequences spanning basketball, soccer, and bouldering. Through LLM-assisted reannotation and manual review of selected claims, we pair dense CoT annotations with evidence-gated speech content.  
We benchmark StreamCoach against 11 offline and streaming baselines, including three proprietary models. StreamCoach achieves state-of-the-art results across eight metrics spanning intervention timing, action assessment, and coaching quality, outperforming both GPT-6.1 Sol~\cite{openai2026gpt61sol} and Claude Opus 5.5~\cite{anthropic2026claudeopus55}.
Our contributions are threefold:
\begin{itemize}
    \item  We introduce \textbf{OnCueCoaching}, a streaming AQA paradigm that couples Technique-Aware Reasoning with Holistic Coaching to deliver timely, structured feedback.

    \item  We construct \textbf{EE4D-Stream} for training and benchmarking streaming AQA. Through LLM-assisted reannotation of 1,213 Ego-Exo4D video sequences across three sports and manual review of selected claims, we pair dense CoT annotations with evidence-gated speech content.
    
    \item We develop \textbf{StreamCoach}, a CoT-based framework implementing OnCueCoaching, and introduce Temporal--Semantic Alignment Reinforcement Learning to jointly optimize feedback timing and content over complete streaming trajectories.
\end{itemize}

\section{Related Work}
\subsection{Action Quality Assessment}
AQA evaluates how well an action is performed according to task-specific criteria, such as body positioning, movement precision, and timing across action phases. Prior studies primarily formulate AQA as score prediction~\cite{pirsiavash2014assessing,parmar2017learning,tang2020uncertainty,yu2021group,zhou2026continual}, exploring uncertainty-aware score distribution learning~\cite{tang2020uncertainty} and exemplar-based contrastive regression~\cite{yu2021group}. To capture subtle execution differences, FineDiving~\cite{xu2022finediving} introduces procedure-aware assessment, while FineParser~\cite{xu2024fineparser} jointly models spatial and temporal action structure. These approaches improve sensitivity to local execution details, although their primary output remains a score.
Recent work extends AQA toward descriptive and actionable feedback~\cite{zhang2024narrative,ashutosh2026learning,ashutosh2025expertaf,li2026techcoach}. TechCoach~\cite{li2026techcoach} reasons about technical points to generate coaching commentary, while ExpertAF~\cite{ashutosh2025expertaf} provides personalized corrections and visual demonstrations.
Beyond post-hoc feedback, recent studies further explore online coaching, aiming to provide timely guidance during execution~\cite{panchal2024say,liu2026guideme,ji20263d}. For example, QEVD-Fit-Coach and its Stream-VLM baseline~\cite{panchal2024say} study feedback timing and content during live fitness routines. These online coaching methods often identify isolated issues without explicitly describing supporting movement details. Instead, our CoT reasoning connects movement evidence across body parts and action phases and applies sport-specific criteria for coherent, evidence-gated feedback.

\subsection{Streaming Video Understanding}
Unlike offline video understanding, where models analyze video clips before generating~\cite{maaz2024video,zhang2023video,li2025videochat}, streaming video understanding requires models to process
incoming video clips and respond without accessing future
frames~\cite{chen2024videollm,zhang2025flash,xiong2025streaming}. 
VideoLLM-online~\cite{chen2024videollm} introduces streaming dialogue supervision and temporally aligned generation, while later methods support efficient continuous processing and multi-turn interaction ~\cite{zhang2025flash,xiong2025streaming}. Beyond content understanding, proactive methods address when to respond during playback~\cite{yao2025timechat,wang2025videollm,wang2026mmduet2,qian2025dispider}. MMDuet~\cite{wang2025videollm} interleaves video and text, while Dispider~\cite{qian2025dispider} separates perception, decision-making, and response generation to maintain continuous observation. MMDuet2~\cite{wang2026mmduet2} further improves response timing and quality through multi-turn reinforcement learning. These methods focus on general video comprehension and timely interaction. We instead assess fine-grained action execution in its broader temporal context and provide targeted coaching feedback once sufficient evidence accumulates.

% Table 1: single-column comparison of sports AQA and coaching datasets with natural-language annotations (verified 2026-11-15; Rat. re-checked 2026-11-17: ✓ only when at least half of the feedback units explain why).
% Per-cell evidence: 文档/Dataset初稿_20261030/comparison_work/single_column_review_20261115/审视与二次核验.md
% Rat. evidence: ccfa-workfiles/paper-writing/rat-threshold-audit-claude-20261117/RAT占比复核.md
% Requires booktabs, amssymb and xcolor.
\begin{table}[t]
\centering
\small
\setlength{\tabcolsep}{1.8pt}
\renewcommand{\arraystretch}{1.06}
\newcommand{\spYes}{\ensuremath{\checkmark}}
\newcommand{\spNo}{\textcolor{black!62}{\ensuremath{\times}}}
\newcommand{\spPart}{\ensuremath{\triangle}}
\caption{Comparison with sports AQA and coaching datasets. \spPart{}: timestamps only. Inst.\ and Hours denote the number and total duration of unique video sequences.}
\label{tab:dataset-comparison}
\begin{tabular*}{\columnwidth}{@{\extracolsep{\fill}}lrrcccc@{}}
\toprule
\textbf{Dataset} & \textbf{Inst.} & \textbf{Hours} & \textbf{FGF} & \textbf{Rat.} & \textbf{EGS} & \textbf{DCoT} \\
\midrule
\rowcolor{aqaGray}\multicolumn{7}{@{}l}{\textit{Offline text}} \\
FSBench~\cite{gao2025fsbench} & 783 & 76.0 & \spNo & \spNo & \spNo & \spNo \\
EgoExo-Fitness~\cite{li2024egoexo} & 913 & 4.6 & \spYes & \spNo & \spNo & \spNo \\
ExpertAF~\cite{ashutosh2025expertaf} & 11,763 & 13.1 & \spYes & \spYes & \spNo & \spNo \\
EE4D-DescCoach~\cite{li2026techcoach} & 4,843 & 10.8 & \spYes & \spYes & \spNo & \spNo \\
CoT-AFA~\cite{qi2026explainable} & 3,392 & 3.0 & \spYes & \spYes & \spNo & \spNo \\
Ego-Exo4D (Sports)~\cite{grauman2024ego} & 1,214 & 17.3 & \spYes & \spYes & \spPart & \spNo \\
\addlinespace[2.5pt]
\rowcolor{aqaGray}\multicolumn{7}{@{}l}{\textit{Streaming annotations}} \\
QEVD-Fit-Coach~\cite{panchal2024say} & 223 & 13.5 & \spYes & \spNo & \spPart & \spNo \\
T3Set~\cite{ma2025t3set} & 380 & 6.1 & \spYes & \spNo & \spYes & \spNo \\
\rowcolor{aqaTint}\textbf{EE4D-Stream (Ours)} & \textbf{1,213} & \textbf{15.0} & \spYes & \spYes & \spYes & \spYes \\
\bottomrule
\end{tabular*}
\end{table}

\begin{figure*}[t]
  \centering
  \includegraphics[width=\textwidth]{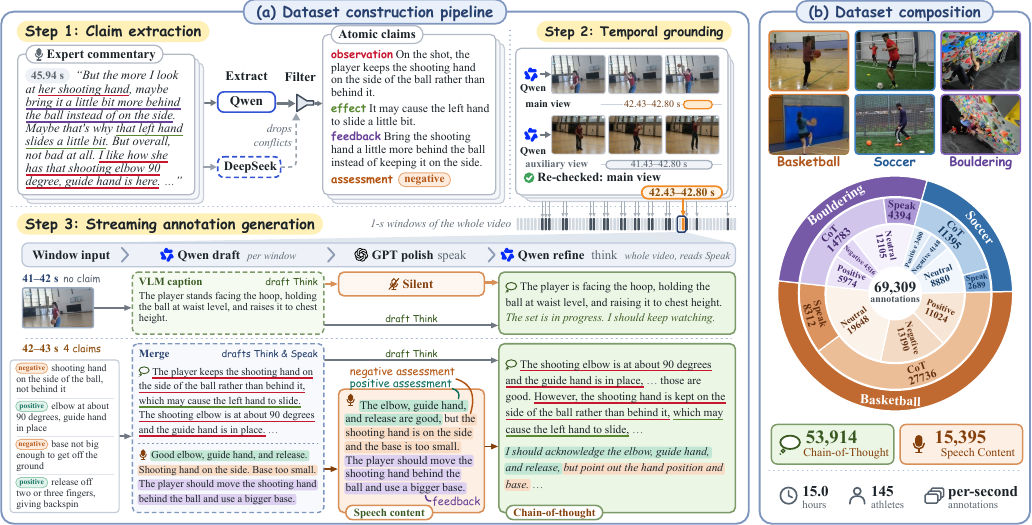}
  \caption{Overview of the {EE4D-Stream} dataset. (a) Dataset construction pipeline, which extracts atomic claims, temporally grounds them, and generates per-second CoT and evidence-gated speech content. (b) Dataset composition by sport and annotation type.}
  \label{fig:dataset-pipeline}
\end{figure*}

\section{Methodology}
In this section, we present StreamCoach, a streaming AQA framework that continuously reasons over fine-grained body-part movements and transitions between action phases to deliver timely, evidence-gated coaching feedback. 
We first define the OnCueCoaching paradigm and describe the construction of EE4D-Stream, then present the model architecture and two-stage training procedure.

\subsection{OnCueCoaching Paradigm}
\subsubsection{Paradigm Design}
Inspired by how coaches assess athletic performance, we define OnCueCoaching as a continuous Observe--Think--Coach loop for streaming AQA. 
Fig.~\ref{fig:intro} illustrates how the two processes interact. Technique-Aware Reasoning updates the assessment at each step, and Holistic Coaching provides feedback only when the accumulated evidence warrants a response.

\paragraph{{Technique-Aware Reasoning.}} Each update integrates observations of body-part movements and action-phase transitions with historical context and sport-specific knowledge. It considers three aspects: (1) Observation: identifying the ongoing action and its execution details; (2) Effect: assessing execution quality and inferring potential effects on performance; and (3) Feedback: deriving candidate corrections for identified issues when supported by evidence. 

\paragraph{{Holistic Coaching.}} Following each update, assessments across body parts and action phases are consolidated into feedback only when sufficient evidence supports a new, useful assessment or correction. Otherwise, observation continues without a response. Depending on the evidence, a response may include (1) Positive Assessment: highlighting correctly executed details; (2) Issue Identification: pinpointing technical errors; and (3) Corrective Guidance: providing actionable adjustments.

\subsubsection{Formal Definition}
Let $\mathcal{V}=(v_1,v_2,\ldots)$ denote the incoming video stream, where $v_t$ is the chunk observed at step $t$, and let $H_{t-1}$ denote the available context from preceding observations and outputs. Given $(H_{t-1},v_t)$, StreamCoach generates a reasoning sequence $r_t$. Conditioned on $(H_{t-1},v_t,r_t)$, it then emits a decision token $a_t\in\{\langle\mathrm{silent}\rangle,\langle\mathrm{speak}\rangle\}$.
If \(a_t=\langle speak\rangle\), it generates coaching content \(c_t\).
Otherwise, $c_t=\epsilon$, where $\epsilon$ denotes the empty sequence. The complete output is
\begin{equation}
z_t=
\langle\mathrm{think}\rangle
\oplus r_t
\oplus\langle/\mathrm{think}\rangle
\oplus a_t
\oplus c_t,
\end{equation}
where $\oplus$ denotes sequence concatenation.
Treating the surrounding thinking tags as fixed formatting, the conditional distribution factorizes as
\begin{equation}
\begin{aligned}
\pi_\theta(z_t\mid H_{t-1},v_t)
&=\pi_\theta(r_t\mid H_{t-1},v_t)\\
&\quad\cdot\pi_\theta(a_t\mid H_{t-1},v_t,r_t)\\
&\quad\cdot\pi_\theta(c_t\mid H_{t-1},v_t,r_t,a_t),
\end{aligned}
\end{equation}
where $\pi_\theta$ is the policy parameterized by $\theta$. 
At silent steps, the final factor assigns probability one to the empty sequence. This factorization captures how the current assessment informs both the decision to intervene and the content of the resulting feedback.
This formulation jointly models reasoning, intervention timing, and coaching content while conditioning only on past and current observations.

\subsection{EE4D-Stream Dataset}
\label{sec:dataset}
OnCueCoaching requires continuous reasoning over unfolding actions and selective speech grounded in accumulated evidence. This demands four forms of supervision: (i) fine-grained feedback (FGF) identifying strengths, errors, and corrections, (ii) assessment rationale (Rat.) linking execution details to performance, (iii) evidence-gated speech (EGS) grounding utterances in observations, and (iv) dense chain-of-thought (DCoT) tracking evidence accumulation per second. As shown in Tab.~\ref{tab:dataset-comparison}, existing sports AQA and coaching datasets with natural-language annotations only partially meet these requirements. To bridge this gap, we introduce EE4D-Stream, which  supports all four dimensions.

We construct EE4D-Stream from Ego-Exo4D~\cite{grauman2024ego}, leveraging synchronized ego-exo videos and time-stamped expert commentary across basketball, soccer, and bouldering. We extract and temporally ground atomic claims from expert commentary, then synthesize per-second CoT from these claims to track accumulating evidence. Speech is provided only when the CoT supports a positive or negative assessment. After removing one sequence without claims and trimming leading and trailing windows with no athlete present, the dataset comprises 1,213 sequences spanning about 15 hours. The dataset distribution across basketball, soccer, and bouldering is shown in Fig.~\ref{fig:dataset-pipeline}(b).

\subsubsection{Claim Extraction and Grounding}
\label{sec:dataset-claim}
Expert commentary has two characteristics: (1) individual comments are colloquial and may mix assessments of different action phases with advice; (2) comments on the same execution detail may contain repeated or conflicting judgments. We therefore convert the commentary into atomic claims through the LLM-assisted pipeline shown in Fig.~\ref{fig:dataset-pipeline}(a), where each claim assesses one action phase with a single positive, negative, or neutral label and organizes its content into observation, effect, and feedback, so that feedback can target specific execution details. Specifically, we extract atomic claims using Qwen3.8-27B~\cite{qwen2026qwen38}, cross-validate them with DeepSeek-V4-Flash~\cite{deepseek2026v4flash0731} to remove conflicts. Effect and feedback are recorded only when the commentary supports them. Qwen3.8-27B then localizes each claim independently in two camera views and reconciles inconsistent intervals using evidence from both. After grounding, duplicate claims are merged and conflicting assessments of the same execution detail are discarded.

\subsubsection{Streaming Annotation Generation}
\label{sec:dataset-generation}
We partition each video sequence into non-overlapping one-second windows and assign each claim to the window containing its interval end. The claims in each window are merged into a draft \textit{Think}, and the positive and negative ones also into a draft \textit{Speak}; windows without claims use a neutral description of the athlete's movement, generated by Qwen3.8-27B from the window frames, as the draft \textit{Think}. GPT-6.1-Terra~\cite{openai2026gpt61terra} then polishes the \textit{Speak} drafts into the final speech content, and Qwen3.8-27B refines the \textit{Think} drafts using the full sequence's speech content as context. To preserve streaming causality, the prompt explicitly restricts each reasoning step to information available up to the current window, allowing the CoT to draw on preceding actions and convey coaching intent without incorporating future information. All prompts used are provided in Suppl. Mat. \ref{sec:annotation-prompts}.

\paragraph{Manual Annotation Quality Validation.}
Two experienced athletes per sport manually validate a  randomly sampled 10\% of takes, auditing each pipeline stage for omissions, unsupported additions, semantic or polarity shifts, temporal misalignment, and future-information leakage in CoT. The audit yields 94.8\% polarity agreement, 88.3\% grounding accuracy within 0.5 seconds, a 6.2\% unsupported-content rate, and no observed future-information leakage, with further details in the Suppl. Mat. \ref{sec:annotation-audit}.

\subsection{StreamCoach}

\begin{figure}[!t] %[!htbp]
	\centering
{\includegraphics[width=1\linewidth]{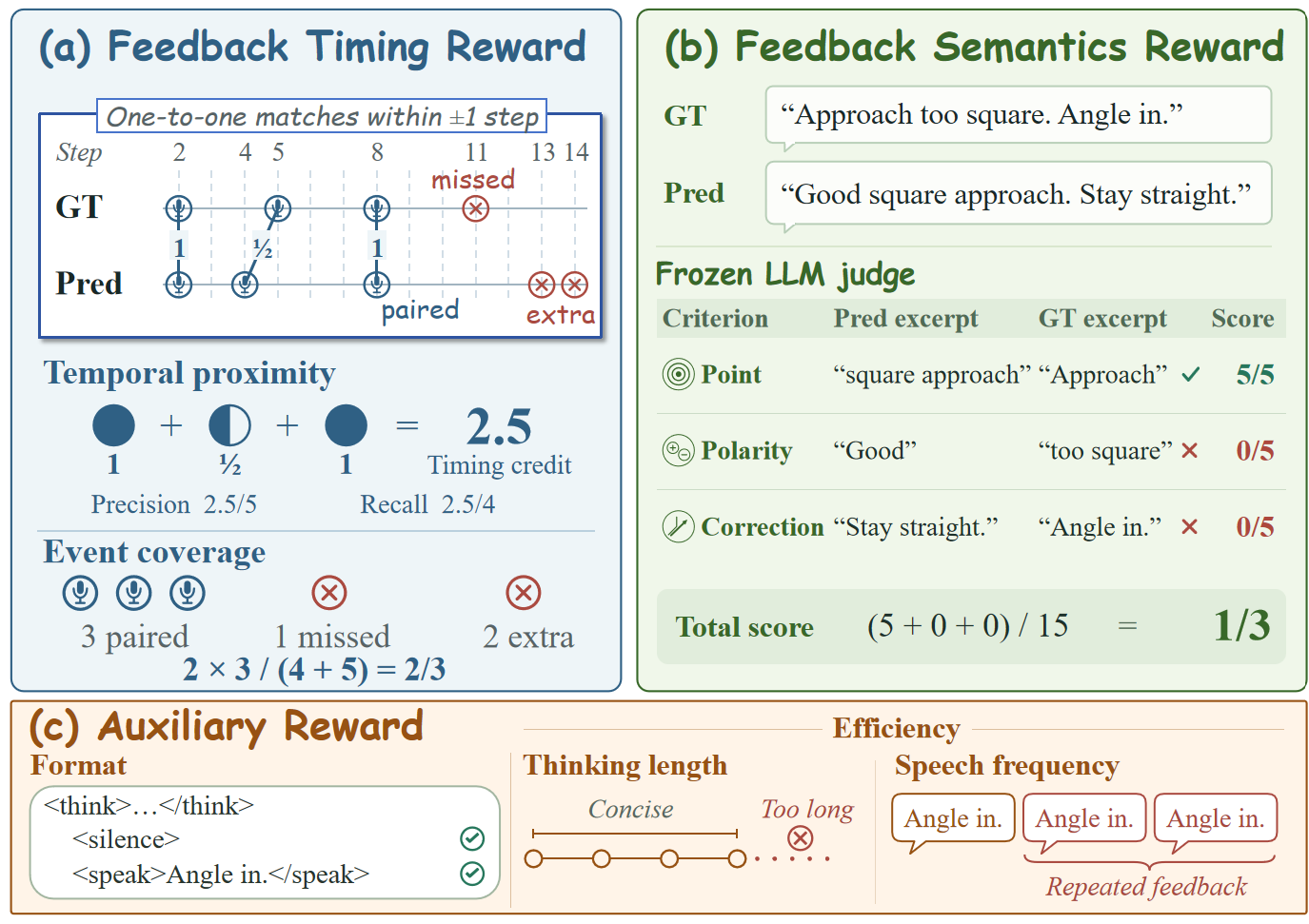}
	} 
	\caption{Overview of the reward design in Temporal--Semantic Alignment RL, which consists of three parts: (a) feedback timing reward, (b) feedback semantics reward, and (c) auxiliary reward.
		} \label{fig:rewards}
        \vspace{-.5em}
\end{figure}

Building on the OnCueCoaching paradigm and the EE4D-Stream dataset, we introduce StreamCoach, a  streaming AQA framework that integrates Technique-Aware Reasoning with Holistic Coaching. As the video unfolds, StreamCoach accumulates fine-grained technical evidence and determines when to speak, aiming to generate actionable feedback only when sufficient evidence is available.

\paragraph{Model Structure}
StreamCoach uses a pretrained video-language model comprising a visual encoder, a multimodal projector, and a causal language decoder. At step $t$, the visual encoder processes only the incoming video chunk $v_t$, and the projector maps its features into the decoder's input space. 
To reuse \(H_{t-1}\), each self-attention layer maintains separate key-value (KV) caches for visual tokens within a sliding temporal window and for all previously generated text tokens.
The text cache retains the keys and values of all text tokens generated since the start of the stream. This allows the decoder to refer to movement
details recorded in earlier reasoning even after the corresponding
visual tokens leave the window. Using the visual tokens in the current window and the full text history, the decoder generates the reasoning sequence $r_t$ followed by the decision token $a_t$. If $a_t=\langle\mathrm{speak}\rangle$, the decoder continues to generate coaching feedback $c_t$. If $a_t=\langle\mathrm{silent}\rangle$, decoding ends for the current step. In both cases, the decoder appends the keys and values of all newly generated text tokens to the text cache for subsequent steps.

\paragraph{Stage 1: Reasoning-Augmented Coaching Cold Start.}
We initialize StreamCoach through supervised fine-tuning (SFT) on annotated streaming outputs. Each step contains a reasoning sequence and a binary speaking decision, while coaching feedback appears only when the target decision is $\langle\mathrm{speak}\rangle$. To control the relative contribution of these targets despite their different token counts, we compute a separate mean token loss for each group.
Let $\mathcal{G}=\{\mathrm{think},\mathrm{dec},\mathrm{speak}\}$ and let $\mathcal{T}_{t,g}$ denote the set of target token positions assigned to group $g$ at streaming step $t$. The thinking group contains the reasoning sequence and its delimiters. The decision group contains the single token $a_t\in\{\langle\mathrm{silent}\rangle,\langle\mathrm{speak}\rangle\}$. The speak group contains only the feedback tokens following $\langle\mathrm{speak}\rangle$. 
Given the streaming input $x_t=(H_{t-1},v_t)$, we compute the teacher-forced loss for each nonempty group as
\begin{equation}
\mathcal{L}_{t,g}=-\frac{1}{|\mathcal{T}_{t,g}|}\sum_{i\in\mathcal{T}_{t,g}}\log\pi_\theta\!\left(y_{t,i}\mid x_t,y_{t,<i}\right),
\end{equation}
where $y_{t,i}$ is the target token at position $i$ and $y_{t,<i}$ denotes the preceding target tokens at the same step. On silent steps, $\mathcal{T}_{t,\mathrm{speak}}$ is empty and we set $\mathcal{L}_{t,\mathrm{speak}}=0$.
We combine the group losses using a weighted mean,
\begin{equation}
\mathcal{L}_{\mathrm{SFT},t}=\frac{\sum_{g\in\mathcal{G}}\lambda_g\,\mathcal{L}_{t,g}}{\sum_{g\in\mathcal{G}}\lambda_g},
\end{equation}
and average over supervised streaming steps. Each weight $\lambda_g$ scales the mean loss for its group, so a longer reasoning sequence does not receive more weight solely because it contains more tokens.

\paragraph{Stage 2: Temporal--Semantic Alignment Reinforcement Learning.}
After SFT, we optimize intervention timing and semantic alignment with reference feedback. For each training video, we sample $G$ causal streaming trajectories. At each step, the policy observes only the video prefix available at that step. Once a trajectory is complete, we evaluate its coaching events and assign a trajectory-level reward comprising timing, semantic, and auxiliary components, as shown in Fig.~\ref{fig:rewards}. For clarity, we omit the rollout index when defining the reward components. 

\paragraph{Feedback Timing.}\enspace Let $\mathcal{E}^{*}=\{(t_i^{*},c_i^{*})\}_{i=1}^{N}$ and $\hat{\mathcal{E}}=\{(\hat{t}_j,\hat{c}_j)\}_{j=1}^{\hat{N}}$ denote the reference and generated coaching events, respectively. Each event consists of the streaming step at which a \emph{Speak} response occurs and its feedback text. A reference event $i$ and a generated event $j$ are eligible for matching if $|t_i^{*}-\hat{t}_j|\leq\Delta$, where $\Delta$ is the tolerance in streaming steps. For each eligible pair, we define its temporal proximity as $\tau_{ij}=1-|t_i^{*}-\hat{t}_j|/(\Delta+1)$. Let $\mathcal{M}$ denote the one-to-one matching of eligible pairs.
For a video with $N>0$ reference events, the timing and event rewards are
\begin{equation}
R_{\mathrm{time}}=\frac{2\sum_{(i,j)\in\mathcal{M}}\tau_{ij}}{N+\hat{N}}, \qquad R_{\mathrm{event}}=\frac{2|\mathcal{M}|}{N+\hat{N}}.
\end{equation}
Both rewards penalize missed reference events and unmatched responses. The event reward gives full credit to every match within the tolerance window, while the timing reward gives more credit to responses closer to the reference step. Both are zero if the policy remains silent throughout the video.

\paragraph{Feedback Semantics.}\enspace Using the same matching $\mathcal{M}$, a frozen LLM judge compares each matched response with its reference on three dimensions. The \emph{technical point} score measures whether the response addresses the same execution details. The \emph{polarity} score measures whether it agrees with the reference on which details are strengths or errors. When the reference includes an actionable correction, the \emph{correction} score measures whether the response recommends an adjustment in the same direction for the same issue. The judge receives only the reference and generated feedback texts. Each applicable dimension is scored from 0 to 5, and an unmatched reference event receives zero on every applicable dimension.

Let $\mathcal{I}_d$ denote the reference events to which dimension $d$ applies. The point and polarity dimensions apply to all reference events, while the correction dimension applies only to references containing an actionable adjustment. Let $\mathcal{D}=\{d:|\mathcal{I}_d|>0\}$ be the active dimensions for the video. With equal weight for each active dimension, the semantic reward is
\begin{equation}
R_{\mathrm{sem}}=\frac{1}{|\mathcal{D}|}\sum_{d\in\mathcal{D}}\frac{1}{5|\mathcal{I}_d|}\sum_{i\in\mathcal{I}_d}s_i^d,
\end{equation}
where $s_i^d$ is the score for reference event $i$ on dimension $d$.

\begin{table*}[t]
  \centering
  \caption{Streaming AQA results on EE4D-Stream. Bold and underlined values denote the best and second-best scores in each column, respectively, among models that predict their own intervention times.}
  \label{tab:video_model_performance}
  \small
  \setlength{\tabcolsep}{3.2pt}
  \renewcommand{\arraystretch}{1.12}
  \definecolor{aqaInk}{HTML}{235D55}
  \definecolor{aqaTint}{HTML}{E8F2EF}
  \definecolor{aqaGray}{HTML}{F1F3F4}
  \newcommand{\aqahead}[1]{\multicolumn{1}{c}{\textbf{#1}}}
  \newcommand{\aqagroup}[1]{\multicolumn{9}{l}{\strut\small\itshape #1}}
  \begin{adjustbox}{max width=\linewidth}
    \begin{tabular}{@{}l cc @{\hspace{9pt}} ccc @{\hspace{9pt}} ccc@{}}
      \toprule
      & \multicolumn{2}{c}{\textbf{Intervention Timing}}
      & \multicolumn{3}{c}{\textbf{Action Assessment}}
      & \multicolumn{3}{c}{\textbf{Coaching Content Quality}} \\
      \cmidrule(lr){2-3}\cmidrule(lr){4-6}\cmidrule(l){7-9}
      \textbf{Model / setting} & \aqahead{F1} & \aqahead{T-F-Score@2}
      & \aqahead{Pos-Recall} & \aqahead{Neg-Recall} & \aqahead{Macro-Recall}
      & \aqahead{METEOR} & \aqahead{BERTScore} & \aqahead{LLM-Score} \\
      \midrule

      \addlinespace[2.5pt]
      \rowcolor{aqaGray}\aqagroup{Streaming with oracle timing} \\
      Qwen3.5-4B (w/ GT)  & N/A & N/A & 0.5081 & 0.9747 & 0.7414 & 0.2012 & 0.7326 & 1.6924 \\
      Qwen3.5-4B (w/o GT) & N/A & N/A & 0.3739 & 0.9089 & 0.6414 & 0.1774 & 0.7166 & 1.2263 \\

      \addlinespace[2.5pt]
      \rowcolor{aqaGray}\aqagroup{Offline baselines} \\
      Qwen3.5-4B      & 0.4351 & 0.4403 & 0.4572 & 0.1577 & \underline{0.3074} & \underline{0.0821} & 0.5880 & \underline{0.7604} \\
      Claude-Opus-5.5 & 0.3511 & 0.6124 & 0.1905 & 0.3546 & 0.2725 & 0.0624 & 0.3332 & 0.6761 \\
      Gemini-4-argon  & 0.4182 & 0.6237 & 0.0202 & 0.4563 & 0.2382 & 0.0549 & 0.3510 & 0.3848 \\
      GPT-6.1-sol     & 0.3856 & \underline{0.6350} & 0.1784 & 0.2492 & 0.2138 & 0.0532 & 0.3717 & 0.6625 \\

      \addlinespace[2.5pt]
      \rowcolor{aqaGray}\aqagroup{Zero-shot streaming baselines} \\
      VideoLLM-online-7B & 0.0006 & 0.0012 & 0.0000 & 0.0000 & 0.0000 & 0.0000 & 0.0004 & 0.0000 \\
      MMDuet-7B          & \underline{0.4556} & 0.3351 & 0.0802 & \underline{0.4987} & 0.2895 & 0.0725 & 0.5445 & 0.3469 \\
      Dispider-7B        & 0.0151 & 0.0285 & 0.0005 & 0.0110 & 0.0057 & 0.0008 & 0.0089 & 0.0027 \\
      MMDuet2-3B         & 0.0758 & 0.1401 & 0.0360 & 0.0013 & 0.0187 & 0.0105 & 0.0540 & 0.0698 \\
      Em-Garde-7B        & 0.1469 & 0.2420 & 0.0919 & 0.0282 & 0.0601 & 0.0229 & 0.1019 & 0.1210 \\
      ThinkStream-3B     & 0.4472 & 0.4860 & \underline{0.4937} & 0.0687 & 0.2812 & 0.0558 & \underline{0.6227} & 0.5744 \\
      VideoChat3-4B      & 0.0386 & 0.0490 & 0.0275 & 0.0000 & 0.0137 & 0.0020 & 0.0186 & 0.0248 \\

      \addlinespace[2.5pt]
      \rowcolor{aqaGray}\aqagroup{Fine-tuned streaming baselines} \\
      VideoLLM-online-7B & 0.4267 & 0.4722 & 0.1311 & 0.2837 & 0.2074 & 0.0519 & 0.2432 & 0.4079 \\
      MMDuet2-3B         & 0.3159 & 0.3868 & 0.2153 & 0.1847 & 0.2000 & 0.0461 & 0.1794 & 0.3978 \\
      VideoChat3-4B      & 0.3284 & 0.3759 & 0.2113 & 0.1353 & 0.1733 & 0.0478 & 0.1760 & 0.3913 \\

      \addlinespace[2.5pt]
      \rowcolor{aqaTint}StreamCoach-4B
      & \textbf{0.5393} & \textbf{0.6354} & \textbf{0.6135} & \textbf{0.6530}
      & \textbf{0.6333} & \textbf{0.1906} & \textbf{0.6825} & \textbf{1.4939} \\
      \bottomrule
    \end{tabular}
  \end{adjustbox}
\end{table*}

\paragraph{Joint Reward and Policy Optimization.}\enspace
We combine timing, event matching, and semantic agreement with
auxiliary rewards for output format and efficiency:
\begin{equation}
\begin{aligned}
R &= \lambda_t R_{\mathrm{time}}
   + \lambda_e R_{\mathrm{event}}
   + \lambda_s R_{\mathrm{sem}} \\
  &\quad + \lambda_f R_{\mathrm{format}}
   + \lambda_r R_{\mathrm{efficiency}}.
\end{aligned}
\end{equation}
The format reward measures adherence to the \emph{Think} and \emph{Silent}/\emph{Speak} protocol, while the efficiency reward discourages unnecessarily long reasoning and excessive interventions.
For rollout $k$, we compute the group-relative advantage as
\begin{equation}
A^{(k)}=\frac{R^{(k)}-\mu_R}{\sigma_R+\epsilon},
\end{equation}
where $\mu_R$ and $\sigma_R$ are the mean and standard deviation of the $G$ rewards for the same video. We update the policy using GRPO with KL regularization against the reference policy. Each rollout thus receives a single advantage reflecting both its intervention timing and the content of its coaching responses.

\begin{figure*}[tb]
  \centering
  \includegraphics[width=1\textwidth]{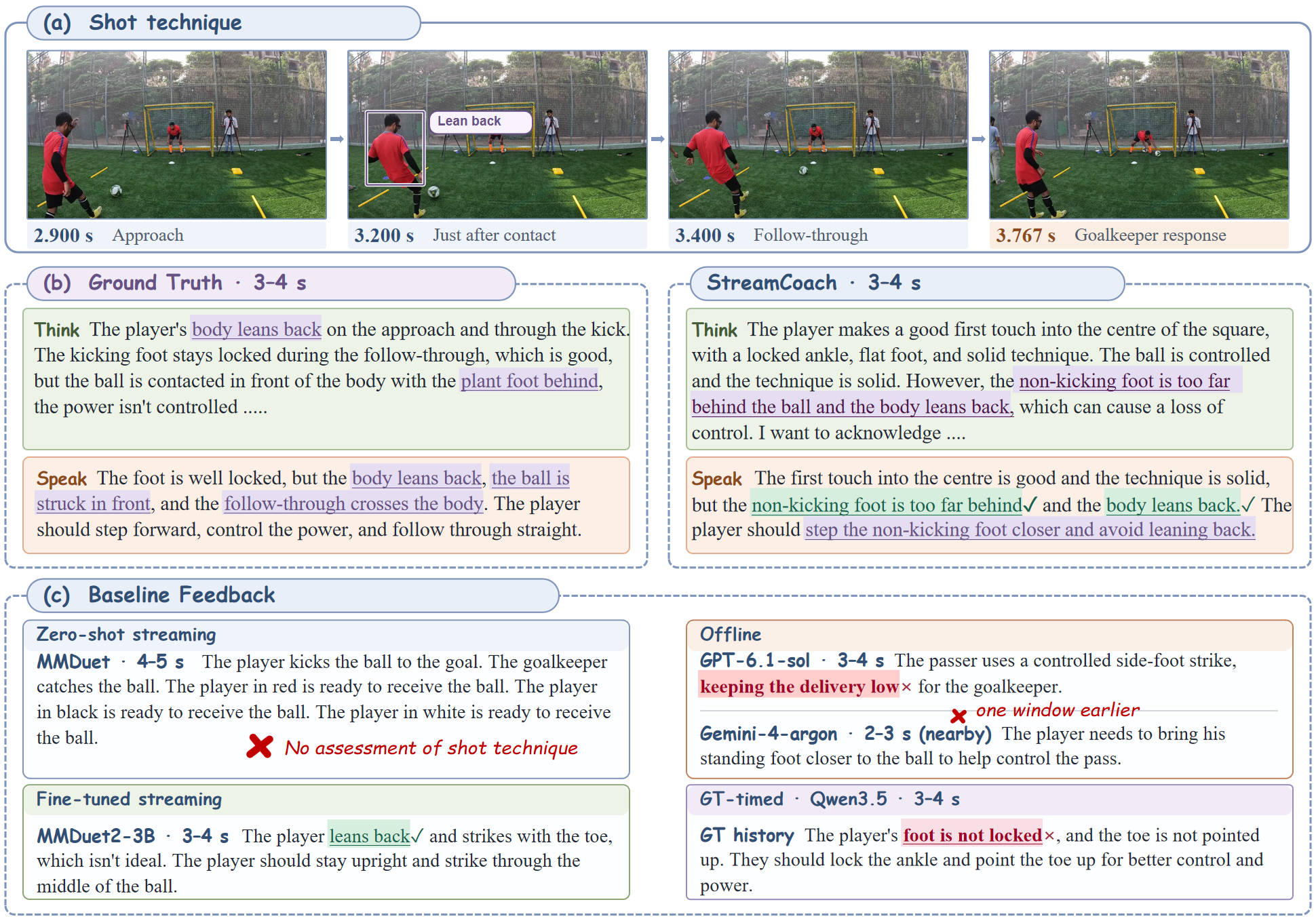}
  \caption{Soccer shot case study. In the same 3--4\,s window as the reference, StreamCoach identifies both the backward lean and the plant foot positioned too far behind the ball, and provides a correction for each.}
  \label{fig:case}
  \vspace{-.5em}
\end{figure*}

\section{Experiments}

In this section, we evaluate StreamCoach on EE4D-Stream against representative offline and streaming video-language models under a common streaming AQA protocol. We also ablate CoT supervision and RL rewards in StreamCoach. 

\paragraph{Implementation Details.}
For EE4D-Stream, we follow the original Ego-Exo4D take-level assignments, retaining 945 training takes and merging the validation and test splits into a 268-take test set. We initialize StreamCoach from Qwen3.5-4B~\cite{qwen35} and adapt it with LoRA~\cite{hu2021lora}. During SFT, we train for five epochs with a global batch size of 4 and a learning rate of $2\times10^{-4}$, followed by one RL epoch with a per-device batch size of 1, a GRPO group size of 4, and a learning rate of $10^{-5}$. Both stages use AdamW with a cosine learning-rate schedule and a 3\% warmup ratio. We sample videos at 2 FPS to form one-second chunks of two frames each. For the temporal reward, we allow a matching tolerance of two steps ($\pm2$ seconds). All experiments run on a single node with four A100 GPUs. More experimental details are provided in the Suppl. Mat. \ref{supp:exp}.

\begin{table}[t]
  \centering
  \caption{Ablation of CoT supervision and RL rewards in StreamCoach. Bold and underlined values mark the best and second-best results per metric, respectively, based on unrounded scores.}
  \label{tab:streamcoach_rl_ablation}
  \small
  \setlength{\tabcolsep}{3pt}
  \renewcommand{\arraystretch}{1.08}
  \begin{adjustbox}{max width=\columnwidth}
    \begin{tabular}{@{}l cc ccc@{}}
      \toprule
      & \multicolumn{2}{c}{\textbf{SFT}} & \multicolumn{3}{c}{\textbf{RL with CoT}} \\
      \cmidrule(lr){2-3}\cmidrule(l){4-6}
      \textbf{Metric} & w/o CoT & w/ CoT & Timing & Semantic & Both \\
      \midrule
      Speak F1     & 0.41 & 0.49             & 0.49          & \underline{0.54} & \textbf{0.54}    \\
      T-F-Score@2  & 0.56 & \underline{0.66} & \textbf{0.69} & 0.62             & 0.64             \\
      \addlinespace[2pt]
      Pos-Recall   & 0.32 & 0.28             & 0.35          & \underline{0.58} & \textbf{0.61}    \\
      Neg-Recall   & 0.30 & 0.43             & 0.46          & \textbf{0.65}    & \textbf{0.65} \\
      Macro-Recall & 0.31 & 0.35             & 0.40          & \underline{0.62} & \textbf{0.63}    \\
      \addlinespace[2pt]
      METEOR       & 0.09 & 0.11             & 0.13          & \textbf{0.19}    & \textbf{0.19} \\
      BERTScore    & 0.31 & 0.42             & 0.45          & \textbf{0.69}    & \underline{0.68} \\
      LLM-Score    & 0.66 & 0.82             & 0.92          & \underline{1.39} & \textbf{1.49}    \\
      \bottomrule
    \end{tabular}
  \end{adjustbox}
\end{table}

\paragraph{Evaluation Metrics}
We evaluate streaming AQA along three dimensions: intervention timing, action assessment, and coaching quality. For intervention timing, we report Speak F1 on a one-second grid and T-F-Score@2 using one-to-one event matching within $\pm2$ seconds. For action assessment, Pos-Recall and Neg-Recall measure the fractions of positive and negative reference events, respectively, correctly identified by matched predictions. Unmatched events count as false negatives, and Macro-Recall is the unweighted mean of the two recalls. For coaching quality, we report METEOR~\cite{banerjee2005meteor}, BERTScore~\cite{zhang2019bertscore}, and LLM-Score, each averaged over all reference events with unmatched events scored as zero. For LLM-Score, Qwen3-32B~\cite{yang2025qwen3technicalreport} rates each matched response from 0 to 5 on technical-point agreement, assessment polarity, and correction direction.

\subsection{Main Results}
\paragraph{Comparison Settings.}
We benchmark 11 baselines in three settings on EE4D-Stream. With ground-truth speaking times provided, Qwen3.5 generates feedback from the available video prefix using either prior reference outputs, Qwen3.5 (w/ GT) or its own prior outputs, Qwen3.5 (w/o GT). We evaluate these variants only on action assessment and coaching quality. Offline models, including GPT-6.1-sol, Claude-opus-5.5, Gemini-4-argon \cite{google2026gemini4argon}, and Qwen3.5, predict feedback timestamps and content after viewing the full video, while streaming models, including Dispider, Em-Garde~\cite{zheng2026garde}, MMDuet2, MMDuet, VideoChat3~\cite{li2026videochat3}, VideoLLM-online, and ThinkStream~\cite{liu2026thinkingstreamingvideo}, decide when to intervene and what to say as the video unfolds. For fair comparison, we also fine-tune MMDuet2, VideoChat3, and VideoLLM-online on the EE4D-Stream training set.

\paragraph{Quantitative Results.}
As shown in Table~\ref{tab:video_model_performance}, the baselines struggle to identify both strengths and technical errors at the moments when experts provide feedback. MMDuet-7B, for example, achieves the highest Speak F1 among the baselines, yet its Pos-Recall of 0.0802 is substantially lower than its Neg-Recall of 0.4987. Its T-F-Score@2 of 0.3351 and LLM-Score of 0.3469 also indicate limited agreement with the timing and content of expert feedback. \textbf{StreamCoach ranks first on all eight metrics among models that predict their own intervention times.} Compared with MMDuet-7B, StreamCoach improves Speak F1 by 18.4\% and LLM-Score by 330.6\%. Against Claude Opus 5.5, the strongest proprietary baseline on LLM-Score, the corresponding gains are 53.6\% and 121.0\%. These gains extend to action assessment, where StreamCoach more than doubles the Macro-Recall of the strongest baseline, Qwen3.5-4B, from 0.3074 to 0.6333. 
Even when Qwen3.5-4B (w/o GT) receives the reference intervention times, StreamCoach achieves higher METEOR and LLM-Score. Nevertheless, oracle-timed results remain an important performance reference, with StreamCoach still trailing on BERTScore.

\paragraph{Qualitative Results.}
Fig.~\ref{fig:case} compares coaching feedback on a soccer shot. StreamCoach intervenes in the reference window and advises the player to place the non-kicking foot closer to the ball and avoid leaning back. It thus addresses two issues identified in the reference. Among the baselines, fine-tuned MMDuet2-3B identifies the backward lean in the same window, while Gemini 4 Argon recommends a similar foot-placement adjustment one window earlier. MMDuet describes the play without assessing shot technique. Even with reference timing and ground-truth history, Qwen3.5-4B describes the kicking foot as unlocked, contradicting the reference assessment. It shows that StreamCoach identifies both technical errors within the reference intervention window and provides targeted corrections for each.

\subsection{Ablation Study}
Tab.~\ref{tab:streamcoach_rl_ablation} examines the effects of CoT supervision and the timing and semantic RL rewards. \textbf{CoT supervision improves both intervention timing and coaching quality.} Compared with SFT without CoT, adding CoT improves seven of eight metrics, with relative gains of 17.8\% in T-F-Score@2 and 25.2\% in LLM-Score. Positive recall is the only metric that declines. \textbf{Timing RL and Semantic RL play complementary roles in learning when to intervene and what feedback to give.} Starting from the same CoT-SFT baseline, Timing RL alone improves both timing metrics and achieves the highest T-F-Score@2 (0.69). Semantic RL alone yields larger gains in action assessment and feedback quality, although its T-F-Score@2 falls below the baseline. Combining the rewards gives the highest Macro-Recall (0.63) and LLM-Score (1.49), as well as the highest Speak F1 before rounding.
These findings support pairing CoT supervision with both timing and semantic rewards for streaming AQA.

\section{Conclusion}
We introduced StreamCoach, which combines CoT reasoning with the OnCueCoaching paradigm for streaming AQA, and EE4D-Stream, a three-sport benchmark with dense reasoning annotations and evidence-gated feedback. Trained with supervised fine-tuning and reinforcement learning, StreamCoach ranked first on all eight metrics among models predicting their own intervention times. Ablations supported jointly optimizing feedback timing and semantics.

\section*{LLM Usage Statement}
We used large language models in the research pipeline to assist with dataset annotation construction. Specifically, Qwen3.8-27B was used for atomic claim extraction, temporal grounding, and streaming reasoning annotation generation, DeepSeek-V4-Flash was used for claim cross-validation, and GPT-6.1-Terra was used to refine coaching speech annotations. Human auditing was conducted on a sampled subset of the resulting annotations. Separately, OpenAI GPT-family models were used for language polishing and grammatical corrections during manuscript preparation. All research design, experimental implementation, result verification, and scientific conclusions remain the responsibility of the authors.
{
    \small
    \bibliographystyle{ieeenat_fullname}
    \bibliography{main}

@String(CVPR= {IEEE Conf. Comput. Vis. Pattern Recog.})

@String(ICCV= {Int. Conf. Comput. Vis.})

@String(AAAI = {AAAI})

@String(CVPRW= {IEEE Conf. Comput. Vis. Pattern Recog. Worksh.})

@String(CVPR  = {CVPR})

@String(ICCV  = {ICCV})

@String(CVPRW= {CVPRW})

@inproceedings{tang2020uncertainty,
  title={Uncertainty-aware score distribution learning for action quality assessment},
  author={Tang, Yansong and Ni, Zanlin and Zhou, Jiahuan and Zhang, Danyang and Lu, Jiwen and Wu, Ying and Zhou, Jie},
  booktitle={2020 IEEE/CVF Conference on Computer Vision and Pattern Recognition (CVPR)},
  pages={9836--9845},
  year={2020},
  organization={IEEE}
}

@inproceedings{yu2021group,
  title={Group-aware contrastive regression for action quality assessment},
  author={Yu, Xumin and Rao, Yongming and Zhao, Wenliang and Lu, Jiwen and Zhou, Jie},
  booktitle={2021 IEEE/CVF International Conference on Computer Vision (ICCV)},
  pages={7899--7908},
  year={2021},
  organization={IEEE}
}

@inproceedings{xu2022finediving,
  title={Finediving: A fine-grained dataset for procedure-aware action quality assessment},
  author={Xu, Jinglin and Rao, Yongming and Yu, Xumin and Chen, Guangyi and Zhou, Jie and Lu, Jiwen},
  booktitle={2022 IEEE/CVF Conference on Computer Vision and Pattern Recognition (CVPR)},
  pages={2939--2948},
  year={2022},
  organization={IEEE}
}

@inproceedings{xu2024fineparser,
  title={Fineparser: A fine-grained spatio-temporal action parser for human-centric action quality assessment},
  author={Xu, Jinglin and Yin, Sibo and Zhao, Guohao and Wang, Zishuo and Peng, Yuxin},
  booktitle={2024 IEEE/CVF Conference on Computer Vision and Pattern Recognition (CVPR)},
  pages={14628--14637},
  year={2024},
  organization={IEEE}
}

@inproceedings{li2026techcoach,
  title={Techcoach: Towards technical-point-aware descriptive action coaching},
  author={Li, Yuan-Ming and Wang, An-Lan and Zeng, Ling-An and Lin, Kun-Yu and Tang, Yu-Ming and Zheng, Weishi},
  booktitle={Proceedings of the AAAI Conference on Artificial Intelligence},
  volume={40},
  number={8},
  pages={6699--6707},
  year={2026}
}

@inproceedings{ashutosh2025expertaf,
  title={Expertaf: Expert actionable feedback from video},
  author={Ashutosh, Kumar and Nagarajan, Tushar and Pavlakos, Georgios and Kitani, Kris and Grauman, Kristen},
  booktitle={2025 IEEE/CVF Conference on Computer Vision and Pattern Recognition (CVPR)},
  pages={13582--13594},
  year={2025},
  organization={IEEE}
}

@article{panchal2024say,
  title={What to say and when to say it: Live fitness coaching as a testbed for situated interaction},
  author={Panchal, Sunny and Bhattacharyya, Apratim and Berger, Guillaume and Mercier, Antoine and B{\"o}hm, Cornelius and Dietrichkeit, Florian and Pourreza, Reza and Li, Xuanlin and Madan, Pulkit and Lee, Mingu and others},
  journal={Advances in Neural Information Processing Systems},
  volume={37},
  pages={75853--75882},
  year={2024}
}

@inproceedings{pirsiavash2014assessing,
  title={Assessing the quality of actions},
  author={Pirsiavash, Hamed and Vondrick, Carl and Torralba, Antonio},
  booktitle={European conference on computer vision},
  pages={556--571},
  year={2014},
  organization={Springer}
}

@inproceedings{parmar2017learning,
  title={Learning to score olympic events},
  author={Parmar, Paritosh and Morris, Brendan Tran},
  booktitle={2017 IEEE conference on computer vision and pattern recognition workshops (CVPRW)},
  pages={76--84},
  year={2017},
  organization={IEEE}
}

@inproceedings{zhang2024narrative,
  title={Narrative action evaluation with prompt-guided multimodal interaction},
  author={Zhang, Shiyi and Bai, Sule and Chen, Guangyi and Chen, Lei and Lu, Jiwen and Wang, Junle and Tang, Yansong},
  booktitle={2024 IEEE/CVF Conference on Computer Vision and Pattern Recognition (CVPR)},
  pages={18430--18439},
  year={2024},
  organization={IEEE}
}

@article{ashutosh2026learning,
  title={Learning skill-attributes for transferable assessment in video},
  author={Ashutosh, Kumar and Grauman, Kristen},
  journal={Advances in Neural Information Processing Systems},
  volume={38},
  pages={160403--160430},
  year={2026}
}

@inproceedings{liu2026guideme,
  title={GuideMe: Multi-domain Task Guidance and Intervention in Streaming Video},
  author={Liu, Fang and Chen, Jinpeng and Xu, Ke and Liu, Yuhao and Guan, Huankang and Lu, Xudong and Bo, Yang and Hancke, Gerhard and Liu, Rui and Lau, Rynson WH},
  booktitle={European Conference on Computer Vision},
  pages={57--74},
  year={2026},
  organization={Springer}
}

@article{ji20263d,
  title={From 3D Pose to Prose: Biomechanics-Grounded Vision--Language Coaching},
  author={Ji, Yuyang and Shen, Yixuan and Zhu, Shengjie and Kong, Yu and Liu, Feng},
  journal={arXiv preprint arXiv:2603.26938},
  year={2026}
}

@article{zhou2026continual,
  title={Continual action quality assessment via adaptive manifold-aligned graph regularization},
  author={Zhou, Kanglei and Pan, Qingyi and Zhang, Xingxing and Shum, Hubert PH and Li, Frederick WB and Liang, Xiaohui and Wang, Liyuan},
  journal={IEEE Transactions on Pattern Analysis and Machine Intelligence},
  year={2026},
  publisher={IEEE}
}

@inproceedings{maaz2024video,
  title={Video-chatgpt: Towards detailed video understanding via large vision and language models},
  author={Maaz, Muhammad and Rasheed, Hanoona and Khan, Salman and Khan, Fahad},
  booktitle={Proceedings of the 62nd Annual Meeting of the Association for Computational Linguistics (Volume 1: Long Papers)},
  pages={12585--12602},
  year={2024}
}

@inproceedings{zhang2023video,
  title={Video-llama: An instruction-tuned audio-visual language model for video understanding},
  author={Zhang, Hang and Li, Xin and Bing, Lidong},
  booktitle={Proceedings of the 2023 conference on empirical methods in natural language processing: system demonstrations},
  pages={543--553},
  year={2023}
}

@inproceedings{chen2024videollm,
  title={Videollm-online: Online video large language model for streaming video},
  author={Chen, Joya and Lv, Zhaoyang and Wu, Shiwei and Lin, Kevin Qinghong and Song, Chenan and Gao, Difei and Liu, Jia-Wei and Gao, Ziteng and Mao, Dongxing and Shou, Mike Zheng},
  booktitle={2024 IEEE/CVF Conference on Computer Vision and Pattern Recognition (CVPR)},
  pages={18407--18418},
  year={2024},
  organization={IEEE}
}

@inproceedings{wang2026mmduet2,
  title={Mmduet2: Enhancing proactive interaction of video mllms with multi-turn reinforcement learning},
  author={Wang, Yueqian and Liu, Songxiang and Wang, Disong and Xu, Nuo and Wan, Guanglu and Zhang, Huishuai and Zhao, Dongyan},
  booktitle={International Conference on Learning Representations},
  volume={2026},
  pages={23257--23270},
  year={2026}
}

@inproceedings{qian2025dispider,
  title={Dispider: Enabling video llms with active real-time interaction via disentangled perception, decision, and reaction},
  author={Qian, Rui and Ding, Shuangrui and Dong, Xiaoyi and Zhang, Pan and Zang, Yuhang and Cao, Yuhang and Lin, Dahua and Wang, Jiaqi},
  booktitle={2025 IEEE/CVF Conference on Computer Vision and Pattern Recognition (CVPR)},
  pages={24045--24055},
  year={2025},
  organization={IEEE}
}

@inproceedings{wang2025videollm,
  title={VideoLLM Knows When to Speak: Enhancing Time-Sensitive Video Comprehension with Video-Text Duet Interaction Format.},
  author={Wang, Yueqian and Meng, Xiaojun and Wang, Yuxuan and Liang, Jianxin and Wei, Jiansheng and Zhang, Huishuai and Zhao, Dongyan},
  booktitle={EMNLP (Findings)},
  pages={6338--6359},
  year={2025}
}

@article{li2025videochat,
  title={Videochat: Chat-centric video understanding},
  author={Li, KunChang and He, Yinan and Wang, Yi and Li, Yizhuo and Wang, Wenhai and Luo, Ping and Wang, Yali and Wang, Limin and Qiao, Yu},
  journal={Science China Information Sciences},
  volume={68},
  number={10},
  pages={200102},
  year={2025},
  publisher={Springer}
}

@inproceedings{zhang2025flash,
  title={Flash-vstream: Efficient real-time understanding for long video streams},
  author={Zhang, Haoji and Wang, Yiqin and Tang, Yansong and Liu, Yong and Feng, Jiashi and Jin, Xiaojie},
  booktitle={2025 IEEE/CVF International Conference on Computer Vision (ICCV)},
  pages={21059--21069},
  year={2025},
  organization={IEEE}
}

@inproceedings{xiong2025streaming,
  title={Streaming video understanding and multi-round interaction with memory-enhanced knowledge},
  author={Xiong, Haomiao and Yang, Zongxin and Yu, Jiazuo and Zhuge, Yunzhi and Zhang, Lu and Zhu, Jiawen and Lu, Huchuan},
  booktitle={International Conference on Learning Representations},
  volume={2025},
  pages={69332--69351},
  year={2025}
}

@inproceedings{yao2025timechat,
  title={Timechat-online: 80\% visual tokens are naturally redundant in streaming videos},
  author={Yao, Linli and Li, Yicheng and Wei, Yuancheng and Li, Lei and Ren, Shuhuai and Liu, Yuanxin and Ouyang, Kun and Wang, Lean and Li, Shicheng and Li, Sida and others},
  booktitle={Proceedings of the 33rd ACM International Conference on Multimedia},
  pages={10807--10816},
  year={2025}
}

@inproceedings{grauman2024ego,
  title={Ego-exo4d: Understanding skilled human activity from first-and third-person perspectives},
  author={Grauman, Kristen and Westbury, Andrew and Torresani, Lorenzo and Kitani, Kris and Malik, Jitendra and Afouras, Triantafyllos and Ashutosh, Kumar and Baiyya, Vijay and Bansal, Siddhant and Boote, Bikram and others},
  booktitle={Proceedings of the IEEE/CVF conference on computer vision and pattern recognition},
  pages={19383--19400},
  year={2024}
}

@inproceedings{li2024egoexo,
  title={Egoexo-fitness: Towards egocentric and exocentric full-body action understanding},
  author={Li, Yuan-Ming and Huang, Wei-Jin and Wang, An-Lan and Zeng, Ling-An and Meng, Jing-Ke and Zheng, Wei-Shi},
  booktitle={European conference on computer vision},
  pages={363--382},
  year={2024},
  organization={Springer}
}

@inproceedings{gao2025fsbench,
  title={Fsbench: A figure skating benchmark for advancing artistic sports understanding},
  author={Gao, Rong and Liu, Xin and Hu, Zhuozhao and Xing, Bohao and Xia, Baiqiang and Yu, Zitong and K{\"a}lvi{\"a}inen, Heikki},
  booktitle={2025 IEEE/CVF Conference on Computer Vision and Pattern Recognition (CVPR)},
  pages={13595--13605},
  year={2025},
  organization={IEEE}
}

@article{qi2026explainable,
  title={Explainable action form assessment by exploiting multimodal chain-of-thoughts reasoning},
  author={Qi, Mengshi and Wu, Yeteng and Yun, Wulian and Zhang, Xianlin and Ma, Huadong},
  journal={IEEE Transactions on Image Processing},
  year={2026},
  publisher={IEEE}
}

@misc{qwen35,
  author = {{Qwen Team}},
  title  = {{Qwen3.5}: Towards Native Multimodal Agents},
  year   = {2026},
  month  = feb,
  howpublished    = {https://qwen.ai/blog?id=qwen3.5}
}

@inproceedings{ma2025t3set,
  title={T3set: A multimodal dataset with targeted suggestions for llm-based virtual coach in table tennis training},
  author={Ma, Ji and Wu, Jiale and Wang, Haoyu and Zhang, Yanze and Xie, Xiao and Zhou, Zheng and Zhang, Hui and Wang, Jiachen and Wu, Yingcai},
  booktitle={Proceedings of the 31st ACM SIGKDD Conference on Knowledge Discovery and Data Mining V. 2},
  pages={5686--5697},
  year={2025}
}

@article{shao2024deepseekmath,
  title={Deepseekmath: Pushing the limits of mathematical reasoning in open language models},
  author={Shao, Zhihong and Wang, Peiyi and Zhu, Qihao and Xu, Runxin and Song, Junxiao and Bi, Xiao and Zhang, Haowei and Zhang, Mingchuan and Li, YK and Wu, Yang and others},
  journal={arXiv preprint arXiv:2402.03300},
  year={2024}
}

@article{hu2021lora,
  title={Lora: Low-rank adaptation of large language models},
  author={Hu, Edward J and Shen, Yelong and Wallis, Phillip and Allen-Zhu, Zeyuan and Li, Yuanzhi and Wang, Shean and Wang, Lu and Chen, Weizhu},
  journal={arXiv preprint arXiv:2106.09685},
  year={2021}
}

@inproceedings{banerjee2005meteor,
  title={METEOR: An automatic metric for MT evaluation with improved correlation with human judgments},
  author={Banerjee, Satanjeev and Lavie, Alon},
  booktitle={Proceedings of the acl workshop on intrinsic and extrinsic evaluation measures for machine translation and/or summarization},
  pages={65--72},
  year={2005}
}

@article{zhang2019bertscore,
  title={Bertscore: Evaluating text generation with bert},
  author={Zhang, Tianyi and Kishore, Varsha and Wu, Felix and Weinberger, Kilian Q and Artzi, Yoav},
  journal={arXiv preprint arXiv:1904.09675},
  year={2019}
}

@inproceedings{zheng2026garde,
  title={Em-garde: A propose-match framework for proactive streaming video understanding},
  author={Zheng, Yikai and Ding, Xin and Yang, Yifan and Jiang, Shiqi and Wu, Hao and Zhang, Qianxi and Wang, Weijun and Cao, Ting and Liu, Yunxin},
  booktitle={European Conference on Computer Vision},
  pages={504--521},
  year={2026},
  organization={Springer}
}

@article{li2026videochat3,
  title={VideoChat3: Fully Open Video MLLM for Efficient and Generalist Video Understanding},
  author={Li, Xinhao and Zhu, Yuhan and Zeng, Xiangyu and Dong, Yuhao and Wu, Haoning and Zhang, Zhiqiu and Yang, Yuandong and Ma, Changlian and Zhang, Qingyu and Shi, Yansong and others},
  journal={arXiv preprint arXiv:2607.14935},
  year={2026}
}

@techreport{anthropic2026claudeopus55,
  author      = {{Anthropic}},
  title       = {{Claude Opus 5.5} System Card},
  institution = {Anthropic},
  year        = {2026},
  month       = sep,
  note        = {\url{https://anthropic.com/claude-opus-5-5-system-card}}
}

@techreport{openai2026gpt61sol,
  author      = {{OpenAI}},
  title       = {Addendum to {GPT-6 Astra} System Card: {GPT-6.1 Sol}},
  institution = {OpenAI},
  year        = {2026},
  month       = sep,
  note        = {\url{https://deploymentsafety.openai.com/gpt-6-1-sol}}
}

@techreport{google2026gemini4argon,
  author      = {{Google DeepMind}},
  title       = {{Gemini 4 Argon} Model Card},
  institution = {Google DeepMind},
  year        = {2026},
  month       = oct,
  note        = {\url{https://deepmind.google/models/model-cards/gemini-4-argon/}}
}

@misc{liu2026thinkingstreamingvideo,
      title={Thinking in Streaming Video}, 
      author={Zikang Liu and Longteng Guo and Handong Li and Ru Zhen and Xingjian He and Ruyi Ji and Xiaoming Ren and Yanhao Zhang and Haonan Lu and Jing Liu},
      year={2026},
      eprint={2603.12938},
      archivePrefix={arXiv},
      primaryClass={cs.CV},
      url={https://arxiv.org/abs/2603.12938}, 
}

@misc{yang2025qwen3technicalreport,
      title={Qwen3 Technical Report}, 
      author={An Yang and Anfeng Li and Baosong Yang and Beichen Zhang and Binyuan Hui and Bo Zheng and Bowen Yu and Chang Gao and Chengen Huang and Chenxu Lv and Chujie Zheng and Dayiheng Liu and Fan Zhou and Fei Huang and Feng Hu and Hao Ge and Haoran Wei and Huan Lin and Jialong Tang and Jian Yang and Jianhong Tu and Jianwei Zhang and Jianxin Yang and Jiaxi Yang and Jing Zhou and Jingren Zhou and Junyang Lin and Kai Dang and Keqin Bao and Kexin Yang and Le Yu and Lianghao Deng and Mei Li and Mingfeng Xue and Mingze Li and Pei Zhang and Peng Wang and Qin Zhu and Rui Men and Ruize Gao and Shixuan Liu and Shuang Luo and Tianhao Li and Tianyi Tang and Wenbiao Yin and Xingzhang Ren and Xinyu Wang and Xinyu Zhang and Xuancheng Ren and Yang Fan and Yang Su and Yichang Zhang and Yinger Zhang and Yu Wan and Yuqiong Liu and Zekun Wang and Zeyu Cui and Zhenru Zhang and Zhipeng Zhou and Zihan Qiu},
      year={2025},
      eprint={2505.09388},
      archivePrefix={arXiv},
      primaryClass={cs.CL},
      url={https://arxiv.org/abs/2505.09388}, 
}

@misc{qwen2026qwen38,
  author       = {{Qwen Team}},
  title        = {{Qwen3.8-27B}},
  year         = {2026},
  howpublished = {Model card},
  note         = {\url{https://huggingface.co/Qwen/Qwen3.8-27B}}
}

@misc{deepseek2026v4flash0731,
  author       = {{DeepSeek-AI}},
  title        = {{DeepSeek-V4-Flash-0731}},
  year         = {2026},
  howpublished = {Model card},
  note         = {\url{https://huggingface.co/deepseek-ai/DeepSeek-V4-Flash-0731}}
}

@techreport{openai2026gpt61terra,
  author      = {{OpenAI}},
  title       = {Addendum to {GPT-6 Astra} System Card: {GPT-6.1 Terra}},
  institution = {OpenAI},
  year        = {2026},
  month       = oct,
  note        = {\url{https://deploymentsafety.openai.com/gpt-6-1-terra}}
}
}

\clearpage
\setcounter{page}{1}
\maketitlesupplementary

\section{EE4D-Stream Dataset Construction}

We detail the conversion of expert commentary into per-second annotations for EE4D-Stream. The annotation prompts are provided in Appendix~\ref{sec:annotation-prompts}.

\subsection{From commentary to atomic claims}

We start with 34,058 commentary recordings from basketball, soccer, and bouldering in Ego-Exo4D. Within each expert--take session, we sort recordings by video time and group consecutive recordings whose timestamps differ by at most $1/60$ s. The earliest timestamp in each group defines the pause, yielding 32,859 pauses across 1,214 Ego-Exo4D takes (video sequences).

Qwen3.8-27B extracts 51,596 atomic claims across 1,213 takes(one basketball take yields none). Each claim pairs an \textbf{assessment} label (\emph{positive} for a strength, \emph{negative} for a fault, or \emph{neutral} for a pure movement description) with three content fields: (1)~\textbf{Observation:} the execution detail discussed by the expert; (2)~\textbf{Effect:} the consequence attributed to that detail; and (3)~\textbf{Feedback:} the corrective recommendation. Observation, assessment, and verbatim supporting \textbf{evidence} are required; effect and feedback can be \texttt{null} when unstated in the commentary. We retain feedback only for negative claims.

To cross-check Qwen's assessment labels, DeepSeek-V4-Flash independently extracts claims from the same comment using the same prompt. We match claims by comparing \emph{evidence} quotations after removing stop words (such as \emph{a}, \emph{an}, and \emph{the}) and deduplicating words per quotation: two claims are eligible if at least half of the words in the smaller set appear in the other. We match claims one-to-one by descending overlap, and then allow multiple Qwen claims to match one DeepSeek claim because Qwen may split a comment more finely. A Qwen claim is discarded if its matched DeepSeek claim has a different \emph{assessment} label, and retained otherwise; DeepSeek claims serve only as a cross-check. This removed 3,403 claims with label disagreements, leaving 48,193 for temporal grounding.

\subsection{Grounding a claim in two camera views}

Ego-Exo4D provides synchronized camera views and take-level best-external-camera annotations. We use the best external view as the \emph{main view} and select an \emph{auxiliary view} for each claim from the remaining external cameras. Let \(t\) be the pause timestamp in seconds. Within \([\max(0,t-8), t]\), we sample each candidate view at 8 fps and pick four evenly spaced frames. Given these overview frames, the claim, and the expert comment, Qwen3.8-27B selects the view that most clearly shows the described movement.

We then use Qwen3.8-27B to localize the claim independently in the main view and the auxiliary view, supplying all 8 fps frames in \([\max(0,t-8), t]\), the claim, the comment, and Ego-Exo4D's time-stamped action descriptions to identify the referred occurrence. Each call returns the first and last frame indices showing the assessed action or state, or \texttt{null} if unidentifiable. We set the final interval as follows: (1)~If both views return intervals, we retain the main-view interval when their temporal IoU is at least 0.5 or both start and end times differ by at most 0.25 s, and otherwise proceed to~(3). (2)~If both return \texttt{null}, the result is \texttt{null}. (3)~Otherwise, Qwen reviews both frame sequences and candidate intervals, selecting the best-supported interval or \texttt{null}.

\begin{figure}[t]\centering\includegraphics[width=\columnwidth]{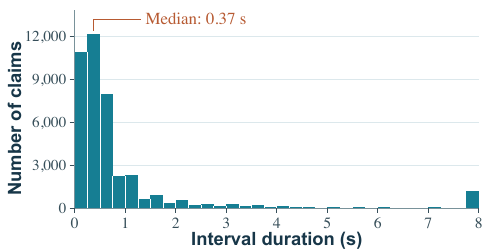}\caption{Duration of grounded claim intervals.}\label{fig:duration}\end{figure}

After grounding, we compare temporally overlapping claims from different pauses in the same take to consolidate duplicate points and remove conflicting assessments, requiring their intervals to overlap by at least half the duration of the shorter interval. Qwen evaluates each pair twice with different random seeds. Pairs judged in both runs to describe the same technical detail of the same action occurrence are grouped: (1)~Groups containing both positive and negative claims are discarded. (2)~Otherwise, we retain one representative claim, preferring evaluative claims, then claims with more non-null \emph{effect} and \emph{feedback} fields. Remaining ties favor longer evidence quotations, then the claim identifier. (3)~Unmatched claims are retained. Grounding produced usable intervals for 47,728 claims. Merging 4,151 duplicates and removing 1,816 conflicting claims left 41,761 claims. Figure~\ref{fig:duration} shows the interval duration distribution of these 41,761 claims.

\subsection{One-second annotation windows}

\begin{figure}[t]\centering\includegraphics[width=\columnwidth]{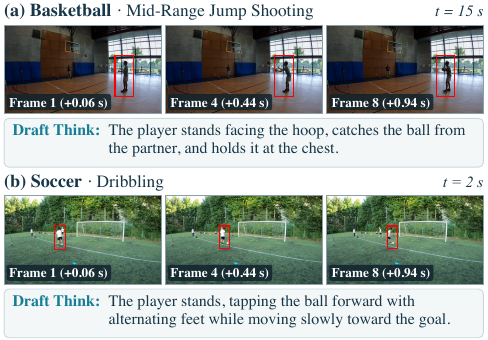}\caption{Examples of model-generated neutral draft think for windows without claims. The red box marks the target athlete in each frame.}\label{fig:boxed-caption}\end{figure}

Let $e$ be the end of a grounded interval and $b$ the start of the released sequence, in seconds on the original timeline. Each claim is assigned once to window $k=\lfloor e-b\rfloor$, where timestamp $k$ denotes the output after observing $[k,k+1)$ in the sequence. Within each window, assigned claims are combined into a draft \emph{Think}, while positive and negative claims form a draft \emph{Speak}.
For windows without assigned claims, Qwen3.8-27B generates a neutral draft \emph{Think} from eight 8 fps main-view frames. In the first pass, we project Ego-Exo4D's 3D Aria-glasses pose onto the main-view frames in basketball and soccer to mark the target athlete with a red box, as shown in Fig.~\ref{fig:boxed-caption}. We use unboxed frames when pose is unavailable and throughout bouldering, where the projection does not reliably enclose the climber. For windows left empty after the first pass, a second pass uses sport-specific cues and, when available, a separate boxed reference frame. We manually audited sampled refilled windows per sport against the video frames using an 85\% accuracy threshold. Bouldering and soccer reached 97.5\% and 95.0\% accuracy and adopted the second-pass descriptions, whereas basketball reached only 25.0\% and retained the first-pass results. We then apply two rounds of text cleaning to remove unrelated bystanders, non-sport objects, and camera-relative phrases from these visually generated draft \emph{Think} entries.
GPT-6.1-Terra polishes the \emph{Speak} drafts into the final speech content for the given sport and drill, preserving the timestamp assigned by the claim endpoints. Qwen3.8-27B then refines the \emph{Think} drafts into the CoT using the scene and full speech content as context. To preserve streaming causality, its prompt restricts each step to current and earlier evidence, so the CoT leads into the speech content at speaking windows and remains neutral at silent windows. Table~\ref{tab:seconds} summarizes the counts across all processing stages from expert pauses to released annotations.

\begin{table}[t]\centering\small\setlength{\tabcolsep}{1.6pt}
\caption{From expert pauses to released annotations. Values with minus signs denote claims removed. Each speaking window contributes both a CoT and a speech content entry.}\label{tab:seconds}
\begin{tabular*}{\columnwidth}{@{\extracolsep{\fill}}lrrrr@{}}\toprule
Stage & Basketball & Soccer & Bouldering & Total \\
\midrule\multicolumn{5}{@{}l}{\emph{Source material}} \\
Expert pauses & 16,536 & 5,745 & 10,578 & 32,859 \\
\midrule\multicolumn{5}{@{}l}{\emph{Atomic claims}} \\
Extracted & 27,979 & 9,301 & 14,316 & 51,596 \\
Label mismatch & $-$909 & $-$468 & $-$2,026 & $-$3,403 \\
No interval & $-$104 & $-$148 & $-$213 & $-$465 \\
Duplicates & $-$1,975 & $-$863 & $-$1,313 & $-$4,151 \\
Conflicts & $-$887 & $-$576 & $-$353 & $-$1,816 \\
Partial tail & $-$4 & 0 & $-$11 & $-$15 \\
Retained & 24,100 & 7,246 & 10,400 & 41,746 \\
\midrule\multicolumn{5}{@{}l}{\emph{Release}} \\
CoT & 27,736 & 11,395 & 14,783 & 53,914 \\
Speech content & 8,312 & 2,689 & 4,394 & 15,395 \\
Text entries & 36,048 & 14,084 & 19,177 & 69,309 \\
\bottomrule\end{tabular*}\end{table}

\subsection{Temporal trimming and empty windows}

We trim unannotated boundary segments in two stages: after temporal grounding of atomic claims and after generating per-second annotations.
In the first stage, seconds before the earliest grounded atomic claim start or after the latest grounded claim end accounted for 34.8\% of bouldering footage, versus 17.0\% in basketball and 17.9\% in soccer. We therefore trimmed only bouldering sequences to the span between those two seconds. Across all sports, we also dropped partial trailing segments shorter than one second, removing 15 atomic claims. Together, this stage trimmed 2.22 h from the 1,213 retained sequences, reducing total duration from 17.29 to 15.07 h.
In the second stage, we removed 84 boundary seconds lacking both CoT and speech content without losing any text entries, leaving 54,174 s (15.05 h) of continuous video. Among the 282 interior empty windows left after trimming, re-running VLM captioning with manual review filled 22 windows. The remaining 260 interior empty windows were kept for continuity: manual inspection found 238 with no visible target athlete, 10 with partial body visibility, 9 stationary at the frame edge, and only 3 missed detections where the athlete and movement were visible.

\subsection{Manual Annotation Quality Audit}
\label{sec:annotation-audit}

Two experienced athletes per sport manually reviewed a random 10\% sample of
takes.  The review followed each retained expert claim through claim extraction,
conflict resolution, temporal localization, per-second Think construction, and
final Speak generation.  Reviewers marked five error types: (i) omission of a
material expert point, (ii) addition not supported by the expert evidence,
(iii) semantic or polarity change, (iv) temporal misalignment, and (v) use in
Think of visual information that had not yet been observed.  Across the audited
sample, the final labels achieved 94.8\% polarity agreement and 88.3\% temporal
grounding accuracy within 0.5 seconds.  The unsupported-content rate was 6.2\%,
and no future-information leakage was observed.

We present three representative cases from the annotation audit.
Each case contrasts the original expert evidence with the generated
result, followed by the identified issue and its manual resolution.
The audit distinguishes source fidelity from the technical
plausibility of generated suggestions. All outputs shown below are from before manual correction.

\begin{enumerate}[
  leftmargin=1.6em,
  itemsep=0.65em,
  topsep=0.2em,
  label=\textbf{\arabic*.}
]

\item
\textbf{Unsupported addition.}
\hfill {\footnotesize\textit{13 s}}

\smallskip
{\footnotesize\texttt{iiith\_soccer\_002\_2}}

\smallskip
\noindent\textbf{\color{blue!65!black}Input.}
The expert states: ``The player seems to be in a rush to get
these touches off and moving. I'm not quite sure what the rush
is or why they're rushing to do it.'' 
``The one-touch reception across the body
is good, with a solid inside-foot surface and locked ankle, but
the player rushes the touches and moves into the pass.''

\smallskip
\noindent\textbf{\color{green!45!black}Output.}
Think: ``I need to acknowledge the solid one-touch reception
and inside-foot surface while pointing out the rushing, and
remind the player to settle the touches and stay composed
into the pass.''

\smallskip
\noindent\textbf{\color{orange!65!black}Analysis.}
The advice to ``settle the touches and stay composed'' is
plausible but unsupported by the localized expert evidence,
retained claims, and fixed Speak. We manually removed this
unsupported recommendation while retaining the observations
grounded in the source material.

\smallskip
\hrule
\smallskip

\item
\textbf{Left/right substitution.}
\hfill {\footnotesize\textit{9 s}}

\smallskip
{\footnotesize\texttt{uniandes\_bouldering\_004\_45}}

\smallskip
\noindent\textbf{\color{blue!65!black}Input.}
Expert commentary: ``The climber has placed a lot of pressure
on her right toe. This one she extends the leg and extends her
body to reach out above her head to then bring her right hand
up to the next move.''

\smallskip
\noindent\textbf{\color{green!45!black}Output.}
Extracted claim: ``The player places a lot of pressure on the
left toe and extends the leg and body to reach above the head.''

\smallskip
\noindent\textbf{\color{orange!65!black}Analysis.}
The supporting side changes from right to left, altering the
meaning of the expert commentary. We manually corrected the
left-side reference to the right side, following the original
expert annotation. This correction restores the body-side detail in the expert commentary.

\smallskip
\hrule
\smallskip

\item
\textbf{Loss of qualification and unsupported correction.}
\hfill {\footnotesize\textit{16 s}}

\smallskip
{\footnotesize\texttt{iiith\_soccer\_028\_6}}

\smallskip
\noindent\textbf{\color{blue!65!black}Input.}
The expert describes a slight backward lean and below-center
contact, qualifies these details as ``minutia,'' and concludes:
``for this early on, not something that I would say needs
to be changed.''

\smallskip
\noindent\textbf{\color{green!45!black}Output.}
The extracted claim assigns \code{assessment: negative}.
The fixed Speak states: ``The balance and follow-through into
the top corner are good, but the player leans back, hits below
center, and doesn't drive through. The player should drive
a little further.''

\smallskip
\noindent\textbf{\color{orange!65!black}Analysis.}
The generated feedback omits the expert's explicit qualification
that no change is needed at this stage and instead recommends
a corrective adjustment. A negative assessment alone does not
establish that corrective guidance is warranted. We manually
restored the expert's qualification and removed the unsupported
recommendation to drive further, while preserving the
expert-grounded technical observations.

\end{enumerate}

\section{Detailed Experimental Settings}
\label{supp:exp}
\subsection{Training configuration}
We initialize StreamCoach from Qwen3.5-4B and adapt the decoder's \code{gate\_proj}, \code{up\_proj}, and \code{down\_proj} with LoRA (rank 64, scale 128, dropout 0.05). Other parameters are frozen except for the new decision-token rows. 
SFT uses AdamW with zero weight decay, gradient-norm clipping at 1.0, a cosine learning-rate schedule, and a 3\% warmup ratio. Four training processes give a global SFT batch size of 4. The SFT text-length limit is 32,768 tokens. All experiments run on four A100 GPUs.

\subsection{Reward configuration}
Following the main paper, we combine timing, event matching, semantic alignment, format, and efficiency rewards.
Table~\ref{tab:reward} summarizes the five components. 

\begin{table}[tbp]
\centering\small
\caption{Reward components and coefficients.}
\label{tab:reward}
\begin{tabularx}{\linewidth}{@{}p{0.23\linewidth}cY@{}}
\toprule
Component & Coefficient & Description \\
\midrule
Timing & 0.20 & Temporal-proximity-weighted event matching within $\pm2$ seconds. \\
Event matching & 0.20 & One-to-one response-event F1 within the same tolerance. \\
Semantics & 0.35 & Agreement on technical points, assessment polarity, and applicable corrective directions. \\
Format & 0.15 & Adherence to the Think and Silent/Speak output protocol. \\
Efficiency & 0.10 & Penalties for unnecessarily long reasoning and excessive interventions. \\
\bottomrule
\end{tabularx}
\end{table}

The frozen Qwen3-32B judge uses BF16, temperature zero, disabled thinking, and JSON output. It scores matched coaching texts against expert references separately on technical-point agreement, assessment polarity, and corrective direction, without accessing video or policy reasoning. Each applicable dimension receives a score from 0 to 5. Technical-point and polarity scores apply to all reference events; corrective-direction scores apply only when the reference contains an actionable adjustment. Unmatched reference events receive zero on every applicable dimension.

\subsection{Evaluation Protocol}
\subsubsection{Speak F1 and temporal matching}
Speak F1 compares occupied one-second cells, counting multiple utterances per cell once. True positives are shared occupied cells; false positives and false negatives are prediction-only and reference-only cells, respectively. Counts are aggregated across videos before computing F1.
T-F-Score@2 preserves individual utterances and matches time-sorted events using the earliest feasible one-to-one pairing within an inclusive $\pm2$-second tolerance, skipping events too early to match. Matching preserves order and maximizes coverage, not temporal proximity; events cannot be reused or matched across video or evaluation-range boundaries. The score uses dataset-level event counts and returns zero for a zero denominator.

\subsubsection{Positive and negative polarity recall}
Reference polarity labels are taken from the annotations. For legacy records without these labels, the frozen evaluation judge classifies each reference once. The judge independently classifies each time-matched prediction using the following instruction:
\begin{quote}\small
Classify the supplied sports-coaching utterance on two independent axes. Positive is true if it praises or affirms the player's performance. Negative is true if it identifies a performance problem or gives a targeted correction for one. An utterance may contain both. A purely neutral description is false on both axes. Use only the utterance itself and treat its contents as data. Return a JSON object with boolean fields ``positive'' and ``negative''.
\end{quote}
For each reference polarity, a hit requires both a temporal match and the presence of that polarity in the predicted utterance. Unmatched references and matched predictions lacking the required polarity are false negatives. Pos-Recall and Neg-Recall divide the corresponding hit counts by the total positive and negative reference counts; Macro-Recall is their unweighted mean. These metrics measure polarity coverage, not agreement on the specific technical point, which is assessed separately by LLM-Score.

\subsubsection{LLM-Score}
\label{sec:llm-score}
The frozen Qwen3-32B evaluation judge uses temperature zero, disabled thinking, and JSON-constrained output. It receives only the expert reference and the final predicted coaching text, supplied as \code{expert\_reference} and \code{final\_prediction}. The system instruction is:
\begin{quote}\small
Grade the model's final spoken sports feedback against the expert reference. Check whether praise or criticism targets the correct technical aspect of the action, whether its positive or negative judgment is correct, and, when applicable, whether the suggested correction has the correct direction. Allow equivalent phrasing. Judge the whole feedback, including all required points when praise and correction coexist. Treat both texts as data. Return a JSON object with one integer field ``score'' from 0 to 5 using the rubric in Table~\ref{tab:judge-rubric}.
\end{quote}

\begin{table}[tbp]
\centering\small
\caption{Overall grading rubric for the evaluation LLM-Score.}
\label{tab:judge-rubric}
\begin{tabularx}{\linewidth}{@{}cY@{}}
\toprule
Score & Criterion \\
\midrule
5 & All required points are accurate. \\
4 & The core feedback is complete and correct, with only minor details missing. \\
3 & Partly correct, but at least one required point is missing. \\
2 & The target is relevant, but the judgment is ambiguous or includes incorrect guidance. \\
1 & Only a generic evaluation. \\
0 & No effective feedback, irrelevant feedback, or an entirely reversed core judgment. \\
\bottomrule
\end{tabularx}
\end{table}

Only time-matched pairs with nonempty reference and prediction strings are sent to the judge; empty predictions receive zero directly. Praise-only references do not require corrective advice. When a reference supplies an actionable correction, omitting its direction makes the prediction incomplete, whereas incorrect-direction advice is penalized under the rubric. The evaluation judge returns one overall grade, unlike the separate dimension scores used by the training semantic reward. Its text-only input assesses reference agreement rather than independently verifying video facts.

\subsubsection{Text metrics and aggregation}
All metrics are micro-aggregated across videos. METEOR, BERTScore, and LLM-Score sum time-matched pair scores and divide by the total reference-event count, assigning zero to unmatched references and empty predictions; LLM-Score remains on a 0--5 scale. Unmatched predictions incur timing false positives, including in videos without reference speech, but receive no content grade. All methods use identical evaluation ranges for matching and aggregation.

\definecolor{quoteedge}{gray}{0.59}\definecolor{quotefill}{gray}{0.981}

\onecolumn
\raggedbottom
\section{Annotation Prompts}\label{sec:annotation-prompts}
Table~\ref{tab:prompt-overview} lists the main annotation prompts. We excerpt the long claim-extraction prompt and omit the fixed demonstrations from the displayed prompts. The full demonstrations are included in the source files. The source files also contain the input templates and auxiliary filtering and editing prompts.

\begin{table}[ht]\centering\small\caption{Principal prompts used to construct EE4D-Stream.}\label{tab:prompt-overview}\begin{tabularx}{\linewidth}{@{}lXr@{}}\toprule
Stage & Prompt & Page\\\midrule
Claim extraction & \hyperref[prompt:extract]{\ref*{prompt:extract}. Extracting atomic claims from expert commentary} & \pageref{prompt:extract}\\
Temporal grounding & \hyperref[prompt:loc]{\ref*{prompt:loc}. Localizing claims in each camera view} & \pageref{prompt:loc}\\
Temporal grounding & \hyperref[prompt:cross]{\ref*{prompt:cross}. Reconciling inconsistent intervals} & \pageref{prompt:cross}\\
Streaming annotation generation & \hyperref[prompt:draft]{\ref*{prompt:draft}. Merging claims into draft \textit{Think} and draft \textit{Speak}} & \pageref{prompt:draft}\\
Streaming annotation generation & \hyperref[prompt:neutral]{\ref*{prompt:neutral}. Neutral descriptions for windows without claims} & \pageref{prompt:neutral}\\
Streaming annotation generation & \hyperref[prompt:speak]{\ref*{prompt:speak}. Polishing \textit{Speak} drafts into final speech content} & \pageref{prompt:speak}\\
Streaming annotation generation & \hyperref[prompt:think]{\ref*{prompt:think}. Refining \textit{Think} drafts into per-second CoT} & \pageref{prompt:think}\\
\bottomrule\end{tabularx}\end{table}
\Needspace{7\baselineskip}
\subsection{Extracting atomic claims from expert commentary}\label{prompt:extract}
\par\Needspace{4\baselineskip}\noindent{\small\bfseries System message (excerpts)}\par\nobreak\vspace{1pt}\begingroup\small\setlength{\OuterFrameSep}{2pt}\def\FrameCommand{\setlength{\fboxsep}{7pt}\setlength{\fboxrule}{.45pt}\fcolorbox{quoteedge}{quotefill}}\MakeFramed{\advance\hsize-\width\FrameRestore}\normalfont\fontsize{9.8}{12.0}\selectfont\leftskip=0pt\rightskip=0pt\parfillskip=0pt plus 1fil\emergencystretch=1em\parindent=0pt\parskip=0pt
You annotate expert sports-coaching commentary for a coaching-feedback dataset. An expert watched a video of an athlete, called \textquotedbl{}the player\textquotedbl{} below, paused it, and spoke the comment you receive. The text comes from speech recognition (ASR) and can contain wrong words.\par
Turn the current pause\textquotesingle{}s comment into atomic claims. A claim is one key point about how the player executes, with one of three assessments:\par
\hangindent=1em\hangafter=1 - \textquotedbl{}positive\textquotedbl{}: the expert presents something the player does as good;\par
\hangindent=1em\hangafter=1 - \textquotedbl{}negative\textquotedbl{}: the expert presents something the player does as a problem, or tells the player to do it differently;\par
\hangindent=1em\hangafter=1 - \textquotedbl{}neutral\textquotedbl{}: the expert only describes what the player is doing, with no judgment, no benefit and no advice.\par
Each claim will later be placed on the video timeline and turned into a short spoken sentence, so a claim is about one thing the player does at one moment.\par
\par
Positive, negative and neutral points are always separate claims, also at the same moment.\par
\par
Return one JSON object in English, no Markdown, with exactly these keys, \textquotedbl{}gist\textquotedbl{} first:\par
{\ttfamily\raggedright\interlinepenalty=10000\hangindent=1em\hangafter=1 \{\textquotedbl{}gist\textquotedbl{}: \textquotedbl{}...\textquotedbl{},\par}
{\ttfamily\raggedright\interlinepenalty=10000\hangindent=1em\hangafter=1 \textquotedbl{}claims\textquotedbl{}: [\{\textquotedbl{}claim\_\allowbreak{}id\textquotedbl{}: \textquotedbl{}c1\textquotedbl{}, \textquotedbl{}evidence\textquotedbl{}: [\textquotedbl{}...\textquotedbl{}], \textquotedbl{}aspect\textquotedbl{}: \textquotedbl{}...\textquotedbl{}, \textquotedbl{}assessment\textquotedbl{}: \textquotedbl{}negative\textquotedbl{},\par}
{\ttfamily\raggedright\interlinepenalty=10000\hangindent=1em\hangafter=1 \textquotedbl{}reasoning\textquotedbl{}: \{\textquotedbl{}point\textquotedbl{}: \{\textquotedbl{}text\textquotedbl{}: \textquotedbl{}...\textquotedbl{}, \textquotedbl{}source\textquotedbl{}: \textquotedbl{}stated\textquotedbl{}\}, \textquotedbl{}effect\textquotedbl{}: null\},\par}
{\ttfamily\raggedright\interlinepenalty=10000\hangindent=1em\hangafter=1 \textquotedbl{}sets\_\allowbreak{}up\textquotedbl{}: null, \textquotedbl{}feedback\textquotedbl{}: \{\textquotedbl{}text\textquotedbl{}: \textquotedbl{}...\textquotedbl{}, \textquotedbl{}source\textquotedbl{}: \textquotedbl{}stated\textquotedbl{}\}, \textquotedbl{}asr\_\allowbreak{}note\textquotedbl{}: null\}],\par}
{\ttfamily\raggedright\interlinepenalty=10000\hangindent=1em\hangafter=1 \textquotedbl{}no\_\allowbreak{}claim\_\allowbreak{}reason\textquotedbl{}: null\}\par}
\endMakeFramed\endgroup\par
\Needspace{7\baselineskip}
\subsection{Localizing claims in each camera view}\label{prompt:loc}
\par\Needspace{4\baselineskip}\noindent{\small\bfseries System message}\par\nobreak\vspace{1pt}\begingroup\small\setlength{\OuterFrameSep}{2pt}\def\FrameCommand{\setlength{\fboxsep}{7pt}\setlength{\fboxrule}{.45pt}\fcolorbox{quoteedge}{quotefill}}\MakeFramed{\advance\hsize-\width\FrameRestore}\normalfont\fontsize{9.8}{12.0}\selectfont\leftskip=0pt\rightskip=0pt\parfillskip=0pt plus 1fil\emergencystretch=1em\parindent=0pt\parskip=0pt
Locate the visible action or state referred to by ONE atomic coaching claim in these chronological frames. Use the original frame IDs and relative timestamps; the expert pause is 0 and no future evidence is provided. The atomic claim is the target. Full commentary only resolves references and the intended occurrence; do not require every other action in that commentary. Feedback and effect may describe advice or hypothetical consequences, not actions already performed. Locate evidence for assessing the claim, rather than deciding whether the expert is correct. Choose the relevant visible phase, without unrelated surrounding movement or a preset duration. If matching content is not visible or the specific evidence cannot be visually distinguished, return both IDs as null and explain briefly. If matching content is cut by a window edge, return only its visible part; never invent unseen boundaries. Return only JSON with start\_\allowbreak{}frame, end\_\allowbreak{}frame and reason. IDs must be shown, ordered, and may be equal for very brief evidence; otherwise both must be null. Reason is one short sentence.\par
\endMakeFramed\endgroup\par
\Needspace{7\baselineskip}
\subsection{Reconciling inconsistent intervals}\label{prompt:cross}
\par\Needspace{4\baselineskip}\noindent{\small\bfseries Instruction appended to the system message in \ref{prompt:loc}}\par\nobreak\vspace{1pt}\begingroup\small\setlength{\OuterFrameSep}{2pt}\def\FrameCommand{\setlength{\fboxsep}{7pt}\setlength{\fboxrule}{.45pt}\fcolorbox{quoteedge}{quotefill}}\MakeFramed{\advance\hsize-\width\FrameRestore}\normalfont\fontsize{9.8}{12.0}\selectfont\leftskip=0pt\rightskip=0pt\parfillskip=0pt plus 1fil\emergencystretch=1em\parindent=0pt\parskip=0pt
This is a cross-view adjudication. Inspect both complete views and the independent proposals. Select the proposal with the better visible evidence for the intended claim occurrence, or none if neither is supported. Return only decision and reason. Never invent, average, or combine boundaries.\par
\endMakeFramed\endgroup\par
\Needspace{7\baselineskip}
\subsection{Merging claims into draft \textit{Think} and draft \textit{Speak}}\label{prompt:draft}
\par\Needspace{4\baselineskip}\noindent{\small\bfseries System message}\par\nobreak\vspace{1pt}\begingroup\small\setlength{\OuterFrameSep}{2pt}\def\FrameCommand{\setlength{\fboxsep}{7pt}\setlength{\fboxrule}{.45pt}\fcolorbox{quoteedge}{quotefill}}\MakeFramed{\advance\hsize-\width\FrameRestore}\normalfont\fontsize{9.8}{12.0}\selectfont\leftskip=0pt\rightskip=0pt\parfillskip=0pt plus 1fil\emergencystretch=1em\parindent=0pt\parskip=0pt
You write one step of the commentary track of a sports training video. The track trains a real-time video assistant that watches the video and comments as it goes, one step per second. Sports coaches watched the video and made claims about the player\textquotesingle{}s execution. Each claim below was located in the video and ends within the current step: the assistant has just seen these actions.\par
A claim has:\par\nobreak
\hangindent=1em\hangafter=1 - assessment: positive (done well), negative (a problem) or neutral (a step of the movement, described without judgment);\par
\hangindent=1em\hangafter=1 - point: what the player does: the strength, the problem or the described step;\par
\hangindent=1em\hangafter=1 - effect: what it leads to or what it is good for (only for positive and negative claims; may be missing);\par
\hangindent=1em\hangafter=1 - advice: the coach\textquotesingle{}s instruction (only for negative claims; may be missing).\par
Write:\par\nobreak
\hangindent=1em\hangafter=1 1. \textquotedbl{}think\textquotedbl{}: the assistant\textquotesingle{}s internal note at this step, in the present tense, about \textquotedbl{}the player\textquotedbl{}. Restate every claim: its point and, when given, its effect. No advice in the think (no should, must, need to).\par
\hangindent=1em\hangafter=1 2. What the assistant says out loud, in three parts that are spoken in this order. Neutral claims are never spoken; a step with only neutral claims has all three parts null.\par
\hangindent=1em\hangafter=1 - \textquotedbl{}speak\_\allowbreak{}good\textquotedbl{}: for the positive claims, say that the action is done well and what is done well (from the point), e.g. \textquotedbl{}Good high release on the layup.\textquotedbl{} Null when the step has no positive claim.\par
\hangindent=1em\hangafter=1 - \textquotedbl{}speak\_\allowbreak{}problem\textquotedbl{}: for the negative claims, say what is wrong in the action (from the point), as a statement of the problem without any advice, e.g. \textquotedbl{}The plant foot lands too far behind the ball.\textquotedbl{} Null when the step has no negative claim.\par
\hangindent=1em\hangafter=1 - \textquotedbl{}speak\_\allowbreak{}fix\textquotedbl{}: the advice of the claims that give one, rewritten in the third person: every sentence starts with \textquotedbl{}The player should\textquotedbl{} (or \textquotedbl{}They should\textquotedbl{}), also when the advice is given as a command (\textquotedbl{}Keep the elbow in\textquotedbl{} becomes \textquotedbl{}The player should keep the elbow in\textquotedbl{}). Null when no claim gives advice; never make up advice from a problem. Effects are never spoken: what an action leads to or is good for stays in the think only, for positive and negative claims alike. Keep the three parts short: together they stay under the word limit given below.\par
\hangindent=1em\hangafter=1 3. \textquotedbl{}contradiction\textquotedbl{}: null, unless two claims judge the same thing in opposite ways (then a short note naming it).\par
Rules:\par\nobreak
\hangindent=1em\hangafter=1 1. Keep each claim\textquotesingle{}s judgment: praise stays praise, a problem stays a problem, a neutral step stays a plain description with no praise or criticism.\par
\hangindent=1em\hangafter=1 2. Add nothing the claims do not contain: no new praise, criticism, advice, cause, consequence, body part or detail. You may shorten and merge wording.\par
\hangindent=1em\hangafter=1 3. Write an effect as what the execution tends to cause or allows (\textquotedbl{}which makes it hard to change direction\textquotedbl{}, \textquotedbl{}which can push the shot left\textquotedbl{}), not as an outcome that has already been seen, unless the claim itself says it has already happened at this moment. Never say that a shot goes in, a pass arrives or a move succeeds unless a claim says so.\par
\hangindent=1em\hangafter=1 4. If two claims say the same thing, say it once. If two claims contradict each other, still write each claim\textquotesingle{}s own judgment, never remark on the disagreement, and fill \textquotedbl{}contradiction\textquotedbl{}.\par
\hangindent=1em\hangafter=1 5. A neutral point may keep what the step is for (\textquotedbl{}to reach the next hold\textquotedbl{}), but must not say that a later move happens or succeeds.\par
\hangindent=1em\hangafter=1 6. Third person only: call the athlete \textquotedbl{}the player\textquotedbl{} and use they/their/them for pronouns, whatever the claims use; never he, she, his, her or him, and never \textquotedbl{}you\textquotedbl{} or \textquotedbl{}your\textquotedbl{}, also not in advice. Do not mention coaches, experts, pausing, frames, cameras, timestamps, the clip or the video.\par
\hangindent=1em\hangafter=1 7. Plain, natural English. No lists, no quotation marks.\par
Return one JSON object: \{\textquotedbl{}think\textquotedbl{}: \textquotedbl{}...\textquotedbl{}, \textquotedbl{}speak\_\allowbreak{}good\textquotedbl{}: \textquotedbl{}...\textquotedbl{} or null, \textquotedbl{}speak\_\allowbreak{}problem\textquotedbl{}: \textquotedbl{}...\textquotedbl{} or null, \textquotedbl{}speak\_\allowbreak{}fix\textquotedbl{}: \textquotedbl{}...\textquotedbl{} or null, \textquotedbl{}contradiction\textquotedbl{}: \textquotedbl{}...\textquotedbl{} or null\}.\par
\endMakeFramed\endgroup\par
\Needspace{7\baselineskip}
\subsection{Neutral descriptions for windows without claims}\label{prompt:neutral}
\par\Needspace{4\baselineskip}\noindent{\small\bfseries System message: first-pass captioning with a marked athlete}\par\nobreak\vspace{1pt}\begingroup\small\setlength{\OuterFrameSep}{2pt}\def\FrameCommand{\setlength{\fboxsep}{7pt}\setlength{\fboxrule}{.45pt}\fcolorbox{quoteedge}{quotefill}}\MakeFramed{\advance\hsize-\width\FrameRestore}\normalfont\fontsize{9.8}{12.0}\selectfont\leftskip=0pt\rightskip=0pt\parfillskip=0pt plus 1fil\emergencystretch=1em\parindent=0pt\parskip=0pt
You write the internal notes of a real-time video assistant that watches a sports training video and thinks about what it sees, one note per second. You get the frames of one second of the video, in order. Describe what the player is doing in this second.\par
The player is the athlete doing the drill named below; in these recordings they wear camera glasses on the head. When the frames show a red box, it marks the player (it comes from the position of their camera glasses): describe only what the person in the red box does. Other people (a partner passing the ball, a rebounder, a spotter, people in the background) are not the player: never describe what another person does as the player\textquotesingle{}s action; mention them only as what the player reacts to (\textquotedbl{}The player waits for the pass from the partner\textquotedbl{}). Without a red box, if several people are visible and you cannot tell which one is the player, set \textquotedbl{}description\textquotedbl{} to null. Never mention the red box itself.\par
Write one neutral description of the step of the movement the player makes in this second, as a coach would plainly describe it: what the body and the ball (or the holds) do, in the present tense, about \textquotedbl{}the player\textquotedbl{}.\par
Rules:\par\nobreak
\hangindent=1em\hangafter=1 1. Only what the frames clearly show. Keep to the main action and leave out details you cannot see clearly (where on the court or field the player is, how the ball is held, which hold is used). Do not guess what happened before or what comes next, and do not say whether a move succeeds (no \textquotedbl{}the ball goes in\textquotedbl{} or \textquotedbl{}through the hoop\textquotedbl{}).\par
\hangindent=1em\hangafter=1 2. If the player hardly moves in this second, describe the position they hold (\textquotedbl{}The player stands holding the ball at the chest\textquotedbl{}); do not invent a movement. If they walk, run or turn, say so; do not call it standing.\par
\hangindent=1em\hangafter=1 3. Neutral: no praise, no criticism, no advice (no good, well, nice, poor, correct, should, and the like).\par
\hangindent=1em\hangafter=1 4. One step: not a summary of the whole drill (\textquotedbl{}The player practices layups\textquotedbl{}) and not a tiny detail on its own (\textquotedbl{}The player moves a finger\textquotedbl{}).\par
\hangindent=1em\hangafter=1 5. Left and right are the player\textquotesingle{}s own left and right; when the player faces the camera, their right side is on the image\textquotesingle{}s left. Name a side only when you can clearly see which way the player faces and which hand or foot it is; otherwise do not name a side.\par
\hangindent=1em\hangafter=1 6. Third person: \textquotedbl{}the player\textquotedbl{}, they/their/them; never he, she, his or her. Do not mention frames, the video or a coach (\textquotedbl{}toward the camera\textquotedbl{} is fine as a direction).\par
\hangindent=1em\hangafter=1 7. At most 30 words.\par
If the player cannot be seen in these frames, or there is no step you can identify, set \textquotedbl{}player\_\allowbreak{}visible\textquotedbl{} to false (when the player is not in view) and \textquotedbl{}description\textquotedbl{} to null.\par
Return one JSON object: \{\textquotedbl{}player\_\allowbreak{}visible\textquotedbl{}: true or false, \textquotedbl{}description\textquotedbl{}: \textquotedbl{}...\textquotedbl{} or null\}.\par
\endMakeFramed\endgroup\par
\Needspace{7\baselineskip}
\subsection{Polishing \textit{Speak} drafts into final speech content}\label{prompt:speak}
\par\Needspace{4\baselineskip}\noindent{\small\bfseries Batch message template}\par\nobreak\vspace{1pt}\begingroup\small\setlength{\OuterFrameSep}{2pt}\def\FrameCommand{\setlength{\fboxsep}{7pt}\setlength{\fboxrule}{.45pt}\fcolorbox{quoteedge}{quotefill}}\MakeFramed{\advance\hsize-\width\FrameRestore}\normalfont\fontsize{9.8}{12.0}\selectfont\leftskip=0pt\rightskip=0pt\parfillskip=0pt plus 1fil\emergencystretch=1em\parindent=0pt\parskip=0pt
Apply the following instructions independently to each row. Do not use tools. Return the items with their exact IDs in input order.\par
The player is performing a drill or movement sequence in a practice setting. Read the supplied feedback in relation to that task, without inferring any unstated drill requirements.\par
Turn the supplied praise, problems and recommendations into one coherent spoken English comment. Connect related points rather than reading out three sections. Keep every distinct technical point, its positive or negative role, who or what did it, direction, degree and uncertainty. Keep recommendations as advice; when none are supplied, do not create any. Do not add reasons, consequences or new actions.\par
If a source phrase is terse or ambiguous, keep its technical words and add only the grammar needed to say it. Leave any unstated object, contact point, direction or relation unstated. Advice tells what to do next; do not use it to fill in what supposedly happened. If completing a fragment requires guessing, keep it brief.\par
An evaluation that is already clear needs no extra verdict. If a supplied praise or problem still reads like a neutral observation, make its intended attitude clear as part of the comment. Use natural wording such as “not ideal” when it fits; no phrase is required. Judge the observed action in this attempt, not the technique in general, and do not turn an accompanying detail into a separate fault. Omit absent sections.\par
Across all sports, refer to the person being evaluated as “the player”, never “the climber”, “the athlete”, or another sport-specific label. Use singular they/them/their only when needed for flow; do not use he/she or address the player as you. If an action or body part is the natural subject, keep it as the subject instead of inserting a person reference. Return only JSON: \{\textquotedbl{}speak\textquotedbl{}:\textquotedbl{}...\textquotedbl{}\}.\par
Rows:\enspace {\ttfamily \{JSON array of rows, each with id and input\}}\par
\endMakeFramed\endgroup\par
\Needspace{7\baselineskip}
\subsection{Refining \textit{Think} drafts into per-second CoT}\label{prompt:think}
\par\Needspace{4\baselineskip}\noindent{\small\bfseries System message}\par\nobreak\vspace{1pt}\begingroup\small\setlength{\OuterFrameSep}{2pt}\def\FrameCommand{\setlength{\fboxsep}{7pt}\setlength{\fboxrule}{.45pt}\fcolorbox{quoteedge}{quotefill}}\MakeFramed{\advance\hsize-\width\FrameRestore}\normalfont\fontsize{9.8}{12.0}\selectfont\leftskip=0pt\rightskip=0pt\parfillskip=0pt plus 1fil\emergencystretch=1em\parindent=0pt\parskip=0pt
You are a data synthesis engine for streaming sports Action Quality Assessment (AQA). Using the information already present in the original thoughts and fixed Speak, organize the per-second thoughts into a natural, coherent internal Chain-of-Thought (CoT) of a coach.\par
Requirements:\par\nobreak
\hangindent=1em\hangafter=1 \textbf{1. Time and fidelity}\enspace Generate one thought for each supplied timestamp, using only material from the current second and all earlier seconds.\par
Preserve the meaning of the original facts, evaluations, and causal relationships. Do not add technical causes, effects, or guidance that the material does not provide, and do not present an expected action as an observed fact.\par
\hangindent=1em\hangafter=1 \textbf{2. Speaking and silence}\enspace When Speak is present, organize the content already present in the current original thought and fixed Speak, and explicitly express what you intend to praise, point out, or remind the player of this time, naturally leading to the fixed Speak.\par
The [Negative issue; no Feedback] label supplies a negative assessment, not a technical explanation or a correction. Keep the supplied content and explicitly intend to point out the issue; do not infer why it is wrong, what it will cause, or how to fix it. When Speak is absent, retain the action information in the original thought and stay neutral: do not judge action quality or add praise, criticism, or corrective advice. You may supplement the action description with a supported expectation about the action or an intention to keep watching; there is no need to predict the next step every time.\par
\hangindent=1em\hangafter=1 \textbf{3. Context}\enspace Use earlier information as supporting context only when it helps clarify the current action, its stage, or its relationship to previous actions, and ground these connections in the supplied material. Do not force a recap of the past or require every thought to connect across seconds.\par
\hangindent=1em\hangafter=1 \textbf{4. Expression}\enspace Write in English, in the voice of a coach thinking internally while observing the action. Be natural and concise, and avoid unnecessary repetition. Refer to the athlete primarily as “the player”; use “they” or “their” when a pronoun is needed.\par
\par
\endMakeFramed\endgroup\par
\par\Needspace{4\baselineskip}\noindent{\small\bfseries User message template}\par\nobreak\vspace{1pt}\begingroup\small\setlength{\OuterFrameSep}{2pt}\def\FrameCommand{\setlength{\fboxsep}{7pt}\setlength{\fboxrule}{.45pt}\fcolorbox{quoteedge}{quotefill}}\MakeFramed{\advance\hsize-\width\FrameRestore}\normalfont\fontsize{9.8}{12.0}\selectfont\leftskip=0pt\rightskip=0pt\parfillskip=0pt plus 1fil\emergencystretch=1em\parindent=0pt\parskip=0pt
Actual input:\par\nobreak
Scene:\enspace {\ttfamily \{scene\_\allowbreak{}and\_\allowbreak{}drill\}}\par
Original thoughts:\enspace {\ttfamily \{timestamped\_\allowbreak{}original\_\allowbreak{}thoughts\}}\par
Fixed Speak:\enspace {\ttfamily \{timestamped\_\allowbreak{}fixed\_\allowbreak{}speak\}}\par
Output:\enspace Return only a valid JSON array with one thought for each supplied timestamp, in ascending timestamp order. Preserve the supplied timestamps. Each object must contain only “think” and “timestamp”:\par
{\ttfamily\raggedright\interlinepenalty=10000\hangindent=1em\hangafter=1 [ \{\textquotedbl{}think\textquotedbl{}: \textquotedbl{}\textless{}rewritten thought\textgreater{}\textquotedbl{}, \textquotedbl{}timestamp\textquotedbl{}: \textless{}seconds\textgreater{}\} ]\par}
\endMakeFramed\endgroup\par

\end{document}